\documentclass{article}
\PassOptionsToPackage{numbers, compress}{natbib}

\usepackage[preprint]{neurips_2024}

\usepackage[utf8]{inputenc} 
\usepackage[T1]{fontenc}    
\usepackage{hyperref}       
\usepackage{url}            
\usepackage{booktabs}       
\usepackage{amsfonts}       
\usepackage{nicefrac}       
\usepackage{microtype}      
\usepackage{xcolor}         

\usepackage{bm}
\usepackage{amsmath}
\usepackage{algorithm}
\usepackage{algpseudocode}
\usepackage{mathtools}
\usepackage{wrapfig}
\usepackage{makecell}
\usepackage{graphicx}
\usepackage{enumitem}
\usepackage{subfig}
\usepackage{multirow}
\usepackage{bbding}
\algnewcommand{\LineComment}[1]{\Statex \(\triangleright\) #1}

\title{Deep Data Consistency: a Fast and Robust Diffusion Model-based Solver for Inverse Problems}

\author{
  Hanyu Chen, Zhixiu Hao, Liying Xiao \\
  Tsinghua University\\
  Beijing, 100084 China \\
  \texttt{hy-chen22@mails.tsinghua.edu.cn, \{haozx,xiaoly\}@tsinghua.edu.cn} \\
}

\begin{document}

\maketitle

\begin{abstract}
Diffusion models have become a successful approach for solving various image inverse problems by providing a powerful diffusion prior. Many studies tried to combine the measurement into diffusion by score function replacement, matrix decomposition, or optimization algorithms, but it is hard to balance the data consistency and realness. The slow sampling speed is also a main obstacle to its wide application. To address the challenges, we propose Deep Data Consistency (DDC) to update the data consistency step with a deep learning model when solving inverse problems with diffusion models. By analyzing existing methods, the variational bound training objective is used to maximize the conditional posterior and reduce its impact on the diffusion process. In comparison with state-of-the-art methods in linear and non-linear tasks, DDC demonstrates its outstanding performance of both similarity and realness metrics in generating high-quality solutions with only 5 inference steps in 0.77 seconds on average. In addition, the robustness of DDC is well illustrated in the experiments across datasets, with large noise and the capacity to solve multiple tasks in only one pre-trained model. \href{https://github.com/Hanyu-Chen373/DeepDataConsistency}{Code.}
\end{abstract}

\section{Introduction}
Diffusion models have shown impressive performance in image generation~\cite{dhariwal2021diffusion, rombach2022high}, providing a powerful prior for solving inverse problems~\cite{song2022pseudoinverse, chung2022diffusion, wang2022zero, kawar2022denoising}, such as super-resolution, inpainting, deblurring, and JPEG restoration. The challenge is how to integrate the diffusion prior perfectly with the measurement. Generally, the approaches can be grouped into several types depending on the data consistency step: (1) replacing score function $\nabla_{\bm{x}_t} \log p (\bm{x}_t)$ with conditional score $\nabla_{\bm{x}_t} \log p (\bm{x}_t|\bm{y})$, whereas its estimation is varied in different solvers like MCG~\cite{chung2022improving}, DPS~\cite{chung2022diffusion}, $\Pi$GDM~\cite{song2022pseudoinverse}, PSLD~\cite{rout2024solving}, etc., (2) decomposing samples into the range-space and the null-space~\cite{wang2022zero}, (3) using diffusion prediction as the initial point for optimization algorithms, such as DDS~\cite{chung2023fast} and ReSample~\cite{song2023solving}.

The common purpose of these solvers is to balance the data consistency (the similarity to the measurement) and the realness (the Gaussian distribution of the diffusion process). Although previously mentioned methods try to address problems from different perspectives, for example, maximum likelihood,  variational approach, and projections on the measurement subspace, they are not well balanced, either deviating from the measurement or suffering from unrealness. Moreover, the slow inference speed impedes the application. These methods require 20 to 1000 diffusion sampling steps, along with additional back-propagation or subspace decomposition steps. Some diffusion bridge-based methods~\cite{liu20232, chung2024direct} attempt to start the sampling process from the measurement to decrease the steps, but it demands a freshly trained diffusion model for a single task, which discards existing models and inevitably takes a lot of time.

\begin{figure}
\includegraphics[width=\textwidth]{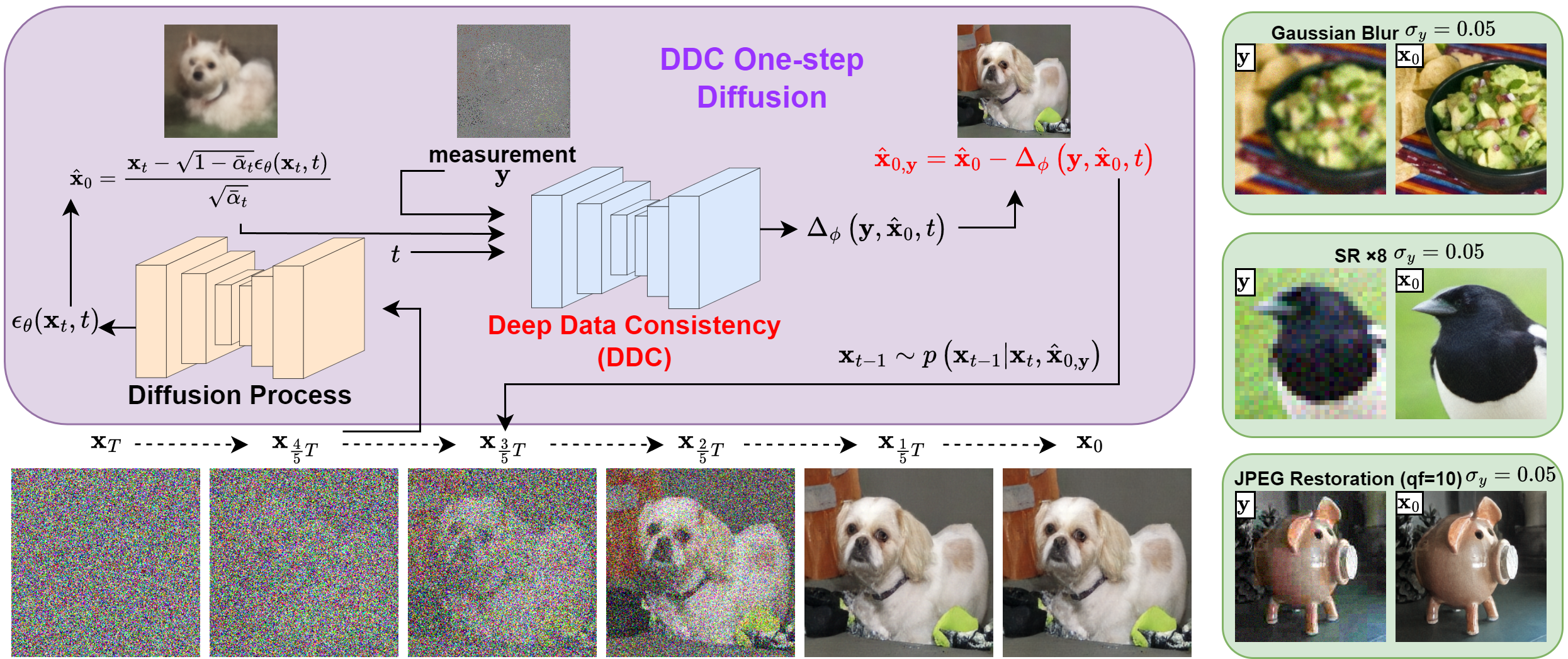}
\caption{DDC solves inverse problems in only \textbf{5 sampling steps}. (\textit{left}) Illustration of our proposed DDC method for a specific random inpainting task. It uses a neural network to update the data consistency step, providing an efficient way for image restoration. (\textit{right}) Some representative results include Gaussian blur, super-resolution $\times8$, and JPEG restoration tasks with large Gaussian noise.}
\label{fig:DDC_illustration}
\end{figure}

To achieve a better trade-off between data consistency and realism, we propose \textbf{D}eep \textbf{D}ata \textbf{C}onsistency (DDC) for solving inverse problems with diffusion models, which uses a neural network to update the data consistency step. The training objective is carefully designed based on variational bound to enable data to approximate real images while minimizing the impact on the diffusion process. In various experiments of both linear and non-linear inverse problems, DDC not only generates high-quality solutions but also proves its rapidity which only requires 5 inference steps (see Fig.~\ref{fig:DDC_illustration}). Moreover, DDC is robust in three aspects: (1) it performs well across datasets, (2) it can work effectively for large noise problems, and (3) multiple tasks can be solved in only one DDC model. 

\paragraph{Contributions.}The main contributions of the paper can be summarized as follows:
\begin{itemize}[noitemsep,nosep]
    \item We propose DDC, a deep-learning-based data consistency solver for inverse problems with diffusion models, which balances data consistency and realness efficiently. The variational training objective of the network is innovatively designed.
    \item DDC generates high-quality and state-of-the-art image solutions in a wide variety of linear and nonlinear inverse problems on ImageNet~\cite{deng2009imagenet} and CelebA~\cite{liu2015faceattributes} datasets.
    \item The DDC framework allows a fast image solution with only 5 sampling steps in 0.77 s. Even though training computation is considered, the accelerated DDC sampling process saves more time overall in comparison.
    \item The robustness of DDC is demonstrated by the fact that one model can learn multiple tasks simultaneously and can easily cope with large noise problems across datasets.
    
\end{itemize}

\section{Background}
\subsection{Denoising Diffusion Models}

Diffusion models~\cite{sohl2015deep,ho2020denoising, song2020score} are generative models that learn the reverse process from Gaussian noise $p_T(\bm{x}) = \mathcal{N} (\bm{0}, \bm{I})$ to the clean data distribution $p_0(\bm{x})$. It corresponds to the forward process that adds noise to the real images $\bm{x}_0 \sim q(\bm{x}_0)$. Once the variance schedule $\beta_1, \beta_2, \dots, \beta_T$ is set, the forward process can be determined as $q(\bm{x}_t) = \mathcal{N} (\sqrt{\bar{\alpha}_t} \bm{x}_0, (1-\bar{\alpha}_t)\bm{I})$, where $\alpha_t = 1 - \beta_t$ and $\bar{\alpha}_t = \prod_{i=1}^t \alpha_i$. The variance preserving (VP) stochastic differential equation (SDE) of the forward and reverse process~\cite{song2020score} can be written as:
\begin{equation}
    d\bm{x}_t = -\frac{\beta_t}{2} \bm{x}_t dt + \sqrt{\beta_t}d\bm{w}
\end{equation}
\begin{equation}
    d\bm{x}_t = \left[-\frac{\beta_t}{2} \bm{x}_t - \beta_t \nabla_{\bm{x}_t} \log p (\bm{x}_t)\right] dt+ \sqrt{\beta_t}d\bar{\bm{w}},
\end{equation}
where $\bm{w}$ and $\bar{\bm{w}}$ are the standard and backward Wiener process, respectively. $\nabla_{\bm{x}_t} \log p (\bm{x}_t)$ is the score function, which is approximated by the score network $\bm{s}_\theta (\bm{x}_t, t) \approx \nabla_{\bm{x}_t} \log p (\bm{x}_t)$ or the noise prediction network $\bm{\epsilon}_\theta (\bm{x}_t, t) \approx -\sqrt{1 - \alpha_t} \nabla_{\bm{x}_t} \log p (\bm{x}_t)$~\cite{dhariwal2021diffusion}. The noise prediction network can be trained by the variant of the variational bound~\cite{ho2020denoising}:
\begin{equation}
    L_\theta = \mathbb{E}_{t, \bm{x}_0, \bm{\epsilon}} \left[ \Vert \bm{\epsilon} - \bm{\epsilon}_\theta (\bm{x}_t, t) \Vert^2 \right],
\end{equation}
where $\bm{\epsilon} \sim \mathcal{N}(\bm{0}, \bm{I})$ is Gaussian noise, and $t$ is uniform between 1 and $T$.

\subsection{Solving Inverse Problems with Diffusion Models
}
\label{sec:inverse_with_diffusion}
Inverse problems can typically be written as:
\begin{equation}
    \bm{y} = \mathcal{A}(\bm{x})+\bm{e}, \enspace \bm{e} \sim \mathcal{N}(\bm{0}, \sigma_{y} \bm{I})
\end{equation}
where $\bm{y} \in \mathbb{R}^m$ is the measurement, $\bm{x} \in \mathbb{R}^n$ is the unknown signal, $\bm{e} \in \mathbb{R}^m$ is the measurement noise, and $\mathcal{A} : \mathbb{R}^m \rightarrow \mathbb{R}^n$ is the forward function. Some solvers simplify it to a linear matrix $\bm{A}$. They are usually ill-posed tasks that require a conditional prior to seeking a reasonable solution. 

When combined with diffusion models, the commonly used method is replacing $\nabla_{\bm{x}_t} \log p (\bm{x}_t)$ with $\nabla_{\bm{x}_t} \log p (\bm{x}_t | \bm{y})$. Following Bayes' rule, it can be modified into two terms~\cite{chung2022improving}:
\begin{equation}
    \nabla_{\bm{x}_t} \log p (\bm{x}_t | \bm{y}) = \nabla_{\bm{x}_t} \log p (\bm{x}_t) + \nabla_{\bm{x}_t} \log p (\bm{y} | \bm{x}_t),
\label{eq:rewrite_score}
\end{equation}
where the network $\bm{\epsilon}_\theta$ estimates the first term $\nabla_{\bm{x}_t} \log p (\bm{x}_t)$, but the analytical formulation of the likelihood $\nabla_{\bm{x}_t} \log p (\bm{y} | \bm{x}_t)$ is hard to acquire. Many previous studies~\cite{kadkhodaie2021stochastic, chung2022improving, rout2024solving} have explored its approximation in depth. For example, DPS~\cite{chung2022diffusion} utilizes some mild assumptions to approximate $\nabla_{\bm{x}_t} \log p (\bm{y} | \bm{x}_t) \simeq \nabla_{\bm{x}_t} \log p (\bm{y} | \bm{x}_0) = \zeta_t \nabla_{\bm{x}_t} \Vert \bm{y} - \mathcal{A}(\hat{ \bm{x}}_0) \Vert^2$, where $\hat{\bm{x}}_0$ is given by Tweedie's formula~\cite{robbins1992empirical} $\hat{ \bm{x}}_0 = \mathbb{E} [\bm{x}_0 | \bm{x}_t] = \frac{1}{\sqrt{\bar \alpha_t}} \left( \bm{x}_t + (1 - {\bar \alpha_t})\nabla_{\bm{x}_t} \log p (\bm{x}_t) \right)$. $\Pi$GDM~\cite{song2022pseudoinverse} assumes that the distribution of $\bm y$ is Gaussian, and the log likelihood then follows $\nabla_{\bm{x}_t} \log p (\bm{y} | \bm{x}_t) = \left( \left( \bm{y} - \bm{A}\hat{ \bm{x}}_0  \right)^\top \left( r_t^2 \bm{A}\bm{A}^\top + \sigma_y \bm{I} \right)^{-1} \bm{A} \frac{\partial\hat{ \bm{x}}_0}{\partial \bm{x}_t} \right)^\top$.

Range-null space decomposition is another useful approach for diffusion image restoration~\cite{wang2022zero}, which rectifies the estimation of $\hat{\bm{x}}_0$ given the observation $\bm{y}$ by $\hat{\bm{x}}_{0,\bm{y}} = \hat{\bm{x}}_0 - \bm{\Sigma}_t \bm{A}^{\dagger} \left( \bm{A} \hat{\bm{x}}_0 - \bm{y} \right)$, where $\bm{\Sigma}_t$ is scale factor, $\bm{A}^{\dagger}$ is the pseudo-inverse, and $\hat{\bm{x}}_0$ is the one-step prediction of $\bm{x}_0$
\begin{equation}
    \hat{\bm{x}}_0 = \frac{1}{{\sqrt {{{\bar \alpha }_t}} }}\left( {{{\bm{x}}_t} - \sqrt {1 - {{\bar \alpha }_t}} {{\epsilon}_\theta }\left( {{{\bm{x}}_t}, t} \right)} \right).
\label{eq: one-step prediction x0}
\end{equation}
Optimization methods, such as the conjugate gradient (CG) and alternating direction method of multipliers (ADMM)~\cite{boyd2011distributed} are also effective in modifying the predicted $\hat{\bm{x}}_0$~\cite{chung2023fast, song2023solving, li2024decoupled}, where the data consistency updates in a more complex way.
\section{Deep Data Consistency}
\label{sec:deep data consistency}
\subsection{Analysis of Existing Methods}

\begin{wrapfigure}[15]{r}{0.4\textwidth}
\vspace{-0.3cm}
\includegraphics[width=0.40\textwidth]{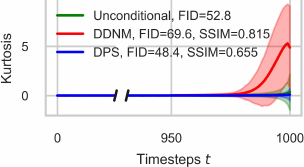}
\caption{The kurtosis of $\bm{\epsilon}_\theta (\bm{x}_t, t)$ when generating images in DPS~\cite{chung2022diffusion}, DDNM~\cite{wang2022zero} and unconditional generation. Fréchet Inception Distance~\cite{heusel2017gans} (FID) and structural similarity (SSIM) are reported to measure the realness and similarity.}
\label{fig:kurtosis}
\end{wrapfigure}

\paragraph{Commonalities}
Although the methods mentioned in Section~\ref{sec:inverse_with_diffusion} vary a lot in their conditional posterior, most of their mathematical forms can be unified as $ \bm{x}_{t, \bm{y}} = \bm{x}_t - \Delta$ or $\hat{\bm{x}}_{0, \bm{y}} = \hat{\bm{x}}_0 - \Delta$, where $\bm{x}_t$ is the noisy image predicted by the reverse process, $\hat{\bm{x}}_0$ is the diffusion approximation of $\bm{x}_0$ at timestep $t$, and $\Delta$ is a function related to $\bm{y}$, $\bm{x}_t$ (or $\hat{\bm{x}}_0$), $\mathcal{A}$, $t$ and $\beta_t$. Detailed discussions are in Appendix~\ref{sec:appendix_commonalities}.

\paragraph{Impact of Data Consistency}
Many approaches have been implemented to reduce the impact of data consistency on the diffusion process, but they do not measure the magnitude of the influence. Since the noisy image $\bm{x}_t$ is used to predict Gaussian noise $\bm{\epsilon}_t$, one possible way is to test whether the output of noise prediction network $\bm{\epsilon}_\theta (\bm{x}_t, t)$ is Gaussian, which can be reflected by its kurtosis $\mathbb{E} \left[ \left( \frac{x-\mu}{\sigma} \right ) ^4 \right] - 3$. A kurtosis closer to 0 indicates its consistency with a Gaussian distribution. Fig.~\ref{fig:kurtosis} shows the kurtosis of $\bm{\epsilon}_\theta (\bm{x}_t, t)$ using DPS~\cite{chung2022diffusion}, DDNM~\cite{wang2022zero} and unconditional image generation when solving super-resolution (SR) $\times4$ task with $\sigma_y = 0$. Details of experiments are in Sec.~\ref{sec: experiement_setup}. The kurtosis of $\bm{\epsilon}_\theta (\bm{x}_t, t)$ in DPS is much closer to 0 than DDNM, which shows that the influence on its diffusion process is less. The smaller impact leads to more realness (smaller FID), which means that the reverse distribution $p(\bm{x}_t)$ in the sampling process is closer to the forward distribution $q(\bm{x}_t)$ in the training procedure. However, with a low SSIM metric, the data consistency of DPS is low, which shows that DPS is a "less conditional" solver. Therefore, the inverse problem solver should ensure data consistency while maximizing realness to improve the realism of the images.

\subsection{Methods}
\label{sec:ddc_methods}
Inspired by the common form of existing methods, we propose DDC that uses neural networks to update data consistency, which may be more flexible and efficient. We summarize the update form, design a training objective to balance realness and data consistency, and overcome the problems of previous studies. Here, the specific details of DDC sampling and DDC training are explained.
\paragraph{DDC Sampling}
DDC tries to use a neutral network $\phi$ to output $\Delta_{\phi}$ instead of complicated mathematical expressions. The input of the model should be $\bm{y}$, $\hat{\bm{x}}_0$ and $t$ while forward function $\mathcal{A}$ and noise schedule $\beta_t$ are implicitly learned in the training process. With this method, the data consistency steps become:
\begin{equation}
    \hat{\bm{x}}_{0, \bm{y}} = \hat{\bm{x}}_0 - \Delta_{\phi} \left(\bm{y}, \hat{\bm{x}}_0, t \right).
\label{eq:ddc_update}
\end{equation}

We choose to update the consistency in the form of Eq.~\ref{eq:ddc_update} for two reasons: (1) although it is not straightforward for the diffusion process to update $\hat{\bm{x}}_0$ rather than $\bm{x}_t$, the clean image data is easier for the network to understand and be compared with the measurement. (2) The residual images $\Delta$ are learned instead of directly predicting clean images $\hat{\bm{x}}_{0, \bm{y}}$ because it has shown great performance in image restoration tasks~\cite{zhang2017beyond, su2022survey}. The sampling process, using DDC as the data consistency step, is shown in Algorithm~\ref{alg: DDC_sampling} in detail.

\begin{figure}[t]
\begin{minipage}{.43\textwidth}
    \begin{algorithm}[H]
        \small
        \caption{DDC sampling}
        \begin{algorithmic}[1]
        \State \textbf{Initialize:} $\bm{x}_T \sim \mathcal{N}(\bm{0}, \bm{I})$
        \For{$t = T$ to $1$} \do \\
            \State $\hat{\bm{x}}_0 = \left(\bm{x}_t - \sqrt{1 - \bar{\alpha}_t} \bm{\epsilon}_{\theta} (\bm{x}_t, t) \right) / \sqrt{\bar{\alpha}_t}$
            \LineComment{Deep Data Consistency}
            \State $\hat{\bm{x}}_{0, \bm{y}} = \hat{\bm{x}}_0 - \Delta_{\phi} \left(\bm{y}, \hat{\bm{x}}_0, t \right)$ 
            \State $\bm{x}_{t-1} \sim p \left(\bm{x}_{t-1} | \bm{x}_t, \hat{\bm{x}}_{0, \bm{y}} \right)$
        \EndFor
        \State \textbf{Return:} $\bm{x}_0$
        \end{algorithmic}\label{alg: DDC_sampling}
    \end{algorithm}
\end{minipage}
\begin{minipage}{.55\textwidth}
    \begin{algorithm}[H]
            \small
           \caption{DDC training}
            \begin{algorithmic}[1]
            \Repeat
                \State $\bm{x}_0 \sim q(\bm{x}_0)$, $t \sim U[1, T]$, $\bm{\epsilon}_y, \bm{\epsilon}_t \sim \mathcal{N} (\bm{0}, \bm{I})$
                \State $\bm{y} = \mathcal{A}(\bm{x}_0)+ \sigma_y \bm{\epsilon}_y$, $\bm{x}_t = \sqrt{\bar{\alpha}_t} \bm{x}_0 + \sqrt{1 - \bar{\alpha}_t} \bm{\epsilon}_t$
                \State $\bm{\epsilon}_\theta = $ StopGradient($\bm{\epsilon}_\theta (\bm{x}_t, t)$)
                \State $\hat{\bm{x}}_0 = \left(\bm{x}_t - \sqrt{1 - \bar{\alpha}_t} \bm{\epsilon}_\theta \right) / \sqrt{\bar{\alpha}_t}$
                \State $\hat{\bm{x}}_{0, \bm{y}} = \hat{\bm{x}}_0 - \Delta_{\phi} \left(\bm{y}, \hat{\bm{x}}_0, t \right)$
                \State $\bm{x}_{t-1} \sim p \left(\bm{x}_{t-1} | \bm{x}_t, \hat{\bm{x}}_{0, \bm{y}} \right)$
                \State $\nabla_{\phi} \left(w_1 L_{MSE} + w_2 L_{LPIPS} + w_3 L_{KL} \right) $
            \Until converged
            \end{algorithmic}\label{alg: DDC_training}
    \end{algorithm}
\end{minipage}
\end{figure}

\paragraph{DDC Variational Bound Training} As discussed above, the goal of inverse problem solvers is to balance data consistency and realness. To train the network $\phi$, we expect to maximize the conditional posterior $p(\hat{\bm{x}}_{0, \bm{y}}|\bm{y})$ meanwhile minimizing the KL divergence $D_{KL} (q(\bm{x}_t | \bm{x}_0 , \bm{y}) \Vert p(\bm{x}_t | \hat{\bm{x}}_{0, \bm{y}} , \bm{y}))$ for each timestep $t$. Therefore, the training procedure is optimizing the variational bound on negative log-likelihood like similar studies DDPM~\cite{ho2020denoising} and VAE~\cite{odaibo2019tutorial}:
\begin{align}
    \mathbb{E} \left[-\log p(\hat{\bm{x}}_{0, \bm{y}}|\bm{y}) \right] & \leq \mathbb{E} \left[ -\log p(\hat{\bm{x}}_{0, \bm{y}}|\bm{y}) \right] + D_{KL} (q(\bm{x}_t | \bm{x}_0 , \bm{y}) \Vert p(\bm{x}_t | \hat{\bm{x}}_{0, \bm{y}} , \bm{y})) \notag \\
    &= - \mathbb{E} \left [\log p(\hat{\bm{x}}_{0, \bm{y}} | \bm{x}_t , \bm{y}) \right ] + D_{KL} (q(\bm{x}_t | \bm{x}_0 , \bm{y}) \Vert p(\bm{x}_t | \bm{y})).
\label{eq:variational_bound}
\end{align}

The detailed derivation of the formula is in the Appendix~\ref{sec:appendix_variational_bound_detailed}. Since the variational bound is expected to be maximized for all timesteps $t$, the training objective $L_{\phi}$ is written as its expectation:
\begin{equation}
    L_{\phi} = \underbrace{- \mathbb{E}_{\bm{x}_0, t, \bm{x}_t, \bm{y}} \left [\log p(\hat{\bm{x}}_{0, \bm{y}} | \bm{x}_t , \bm{y}) \right ]}_{L_{recon}} + \underbrace{\mathbb{E}_{\bm{x}_0, t, \bm{x}_t, \bm{y}}D_{KL} (q(\bm{x}_t | \bm{x}_0 , \bm{y}) \Vert p(\bm{x}_t | \bm{y}))}_{L_{KL}},
\label{eq:training_objective}
\end{equation}
where the first part $L_{recon} = - \mathbb{E}_{\bm{x}_0, t, \bm{x}_t, \bm{y}} \left [\log p(\hat{\bm{x}}_{0, \bm{y}} | \bm{x}_t, \bm{y}) \right ]$ is the reconstruction loss to measure the likelihood of the data output, while the diffusion model $\theta$ first gives the estimation~\cite{ho2020denoising} by Eq.~\ref{eq: one-step prediction x0}. If we assume that the difference between conditional prediction $\hat{\bm{x}}_{0, \bm{y}}$ and ground truth image $\bm{x}_0$ is i.i.d Gaussian, the term will become mean square error (MSE) loss $L_{recon} = \Vert \hat{\bm{x}}_{0, \bm{y}} - \bm{x}_0 \Vert^2 = \Vert (\hat{\bm{x}}_{0} - \Delta_{\phi} \left(\bm{y}, \hat{\bm{x}}_0, t \right) ) - \bm{x}_0 \Vert^2$. However, recent research points out that models trained with learned perceptual image patch similarity (LPIPS) loss~\cite{zhang2018unreasonable} provide better metric results and more vivid colors~\cite{ballester2022analysis}. Therefore, the LPIPS loss is added to the reconstruction loss term:
\begin{equation}
    L_{recon} = L_{MSE} (\hat{\bm{x}}_{0, \bm{y}}, \bm{x}_0) + L_{LPIPS} (\hat{\bm{x}}_{0, \bm{y}}, \bm{x}_0),
\end{equation}
and we analyze its impact on image restoration in Section~\ref{sec: experiments}.

The second term $L_{KL}$ in Eq.~\ref{eq:training_objective} represents the difference between the distribution of $\bm{x}_t$ in the forward process $q$ and the reverse process $p$ relative to the observation $\bm{y}$. The forward distribution is Gaussian $q(\bm{x}_t | \bm{x}_0 , \bm{y}) = q(\bm{x}_t | \bm{x}_0) = \mathcal{N} (\sqrt{\bar{\alpha}_t} \bm{x}_0, (1-\bar{\alpha}_t)\bm{I})$ because the noisy image $\bm{x}_t$ is only relevant to the clean image $\bm{x}_0$. The reverse process can be chosen as the Gaussian distribution $p(\bm{x}_t | \bm{y}) \sim \mathcal{N} (\bm{\mu}_t, \bm{\sigma}_t)$ so that we can obtain a closed-form loss function, which has been proven efficient in studies~\cite{ho2020denoising, odaibo2019tutorial}. The KL loss then can be written as:
\begin{equation}
    L_{KL} = \frac{1}{2} \left ( \log \frac{\sigma_t^2}{1 - {{\bar \alpha }_t}} + \frac{1 - {{\bar \alpha }_t} + \left(\sqrt {{{\bar \alpha }_t}}x_0 - \mu_t \right)^2}{\sigma_t^2} -1\right ),
\end{equation}
which is further explained in Appendix~\ref{sec:appendix_variational_bound_detailed}. Meanwhile, we can utilize adjustable hyperparameters to balance the loss objectives, leading to the final training objective:
\begin{equation}
    L_{\phi} = w_1 L_{MSE} + w_2 L_{LPIPS} + w_3 L_{KL}.
\label{eq: final_obj}
\end{equation}
With the weighted objective Eq.~\ref{eq: final_obj}, the training process of DDC is shown in Algorithm~\ref{alg: DDC_training}.

\section{Experiments}
\label{sec: experiments}

\subsection{Experiment Setup}
\label{sec: experiement_setup}
\paragraph{Dataset and Problem Setup}For the experiments on image quality, efficiency, and noisy robustness, we compare our proposed DDC with state-of-the-art approaches for solving inverse problems on 256×256 ImageNet-1k~\cite{deng2009imagenet} dataset of 1,000 validation images. The pre-trained diffusion model is provided by~\cite{dhariwal2021diffusion} without fine-tuning. We also use 1000 validation images from the CelebA~\cite{liu2015faceattributes} dataset and the pre-trained diffusion model from~\cite{lugmayr2022repaint} for the robustness test across datasets. Four linear inverse problem tasks are performed as follows: (1) super-resolution (SR) $\times4$ with bicubic downsampling, (2) SR $\times8$ with bicubic downsampling, (3) Gaussian blur with $61\times61$ kernel and standard deviation of 3.0, (4) random inpainting that masks $92\%$ of total pixels on all RGB channels. All the linear tasks are compared with additional Gaussian measurement noise $\sigma_y = 0$ and $0.05$. For the non-linear problem, we choose JPEG restoration with quality factor $QF = 10$ without noise. 

\paragraph{DDC Setup} U-Net~\cite{ronneberger2015u} is selected as the architecture of the data consistency network $\phi$, which is trained with the training dataset of ImageNet-1k~\cite{deng2009imagenet} in only one epoch for the main experiment. Two generalized models (simultaneously trained by all the linear and non-linear tasks) are trained, and the only difference between the two is whether LPIPS loss is added. During training, random Gaussian noise whose standard deviation $\sigma_y$ ranges from 0 to 0.05 is added to the measurement $\bm{y}$. For robustness tests, a DDC model trained on the CelebA~\cite{liu2015faceattributes} dataset, an SR $\times4$ model with a large Gaussian noise range from 0 to 0.25, and a  Gaussian blur task-specific model is also obtained. In the process of DDC sampling, inference steps are set to $T=5$. All the experiments were performed on one NVIDIA GeForce RTX 4090 GPU.

\paragraph{Baselines and Evaluation Metrics} Several diffusion model-based inverse problems are covered, including DPS~\cite{chung2022diffusion}, DDNM~\cite{wang2022zero}, DDRM~\cite{kawar2022denoising}, $\Pi$GDM~\cite{song2022pseudoinverse}, and RED-diff~\cite{mardani2023variational}. PnP-ADMM~\cite{chan2016plug} is also selected as a typical CNN-based method. Moreover, the diffusion bridge-based approach CDDB~\cite{chung2024direct} is included for the non-linear task. The experimental details are shown in Appendix~\ref{sec: appendix_experimental_details}. To evaluate these various methods, we report metrics including peak signal-to-noise ratio (PSNR), SSIM, LPIPS~\cite{zhang2018unreasonable}, and FID~\cite{heusel2017gans}.

\subsection{Results on Quality and Efficiency}
\label{sec:results_main}
\begin{figure}
    \centering
    \begin{minipage}[t]{\textwidth}
    \centering
    \includegraphics[width=\textwidth]{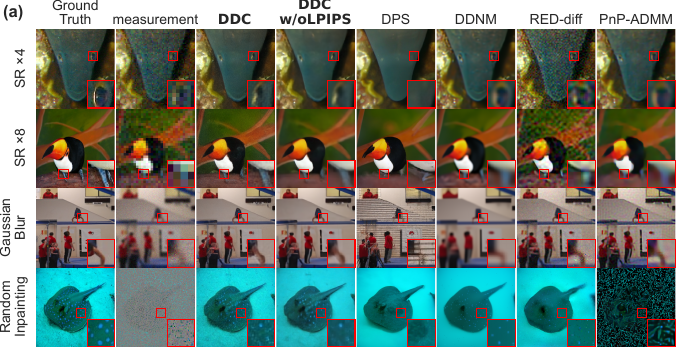}
    \end{minipage}

    \begin{minipage}[t]{\textwidth}
    \centering
    \includegraphics[width=\textwidth]{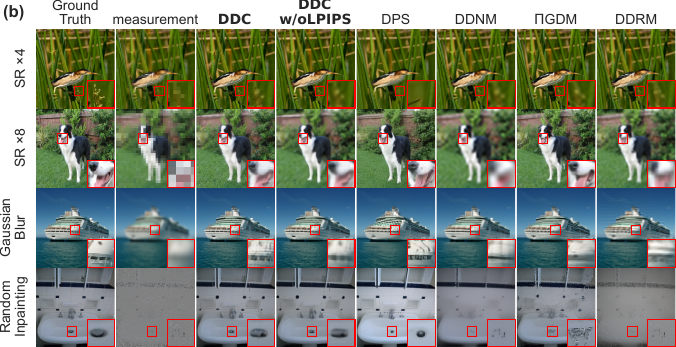}
    \caption{Representative results on linear inverse problems (a) with Gaussian noise $\sigma_y = 0.05$ and (b) without noise on ImageNet dataset. The degradations include SR $\times4$, SR $\times8$, Gaussian blur and random inpainting tasks.}
    \label{fig:representative_results}
    \end{minipage}
    
\end{figure}

\paragraph{Quality Comparison against baselines in linear tasks} DDC outperforms the baselines in all the inverse problem tasks on the ImageNet dataset, as shown in Tab.~\ref{tab:quantitative_sigma005} and~\ref{tab:quantitative_sigma0}. The standard version of DDC (with LPIPS loss) achieves most of the best results of FID and LPIPS metrics, which shows the best realism and human perceptual similarity among the solvers. Although DDC without LPIPS loss has better performance in PSNR and SSIM metrics, it cannot reconstruct many nuances for human perception, leading to more blurry images shown in Fig.~\ref{fig:representative_results}. Therefore, adding LPIPS loss is necessary because it improves realism and better reflects human perception. DPS has good performance in FID, which can be reflected by its excellent realness and less influence on the diffusion process, but it has less similarity to the measurement with poor PSNR and SSIM. It often generates images that are closely related in image topics but exhibit significant differences in specific details (see Fig.~\ref{fig:representative_results}). DDNM has some second-best results on SR tasks in PSNR and SSIM, but it comes at the expense of image realism and generates vague results. DDRM is mediocre on SR and Gaussian blur tasks, but cannot solve random inpainting well. $\Pi$GDM and RED-diff are quite satisfying to solve SR tasks without noise, but they are vulnerable to noise and complex problems like 92$\%$ random inpainting. PnP-ADMM, as a traditional CNN method, is inferior to diffusion model-based solvers in terms of fidelity and data consistency.

\begin{table}[!t]
    \centering
    \setlength{\tabcolsep}{0.2em}
    \caption{Quantitative results of noisy linear tasks including SR $\times4$, SR $\times8$, Gaussian blur, and random inpainting with Gaussian noise $\sigma_y = 0.05$ on the ImageNet dataset. The best results are \textbf{bold}, and the second best results are \underline{underlined}.}
    \resizebox{1.0\textwidth}{!}{
    \begin{tabular}{ccccccccccccccccc}
    \toprule 
{} & \multicolumn{4}{c}{\textbf{SR $\times4$}} & \multicolumn{4}{c}{\textbf{SR $\times8$}} &
\multicolumn{4}{c}{\textbf{Gaussian Blur}} &\multicolumn{4}{c}{\textbf{Random Inpainting}} \\
\cmidrule(lr){2-5}
\cmidrule(lr){6-9}
\cmidrule(lr){10-13}
\cmidrule(lr){14-17}
        {\textbf{Method}} & {PSNR $\uparrow$} & {SSIM $\uparrow$} & {LPIPS $\downarrow$} & {FID $\downarrow$}& {PSNR $\uparrow$} & {SSIM $\uparrow$} & {LPIPS $\downarrow$} & {FID $\downarrow$}& {PSNR $\uparrow$} & {SSIM $\uparrow$} & {LPIPS $\downarrow$} & {FID $\downarrow$}& {PSNR $\uparrow$} & {SSIM $\uparrow$} & {LPIPS $\downarrow$} & {FID $\downarrow$} \\ \midrule
        \textbf{DDC} & 23.77 & 0.742 & \textbf{0.216} & \underline{50.07} & 21.17 & 0.620 & \textbf{0.370} & \underline{65.11} & \underline{23.88} & \underline{0.741} & \textbf{0.206} & \textbf{49.64} & \underline{22.54} & \underline{0.723} & \textbf{0.203} & \underline{51.48} \\ 
        \textbf{DDC w/o LPIPS} & \underline{24.49} & \underline{0.778} & 0.345 & 58.50 & \underline{21.72} & \underline{0.664} & 0.518 & 76.37 & \textbf{24.65} & \textbf{0.776} & \underline{0.355} & 58.12 & \textbf{22.91} & \textbf{0.742} & \underline{0.294} & 60.19 \\ \midrule
        DPS~\cite{chung2022diffusion} & 23.04 & 0.596 & \underline{0.314} & \textbf{47.62} & 20.07 & 0.444 & \underline{0.381} & \textbf{50.09} & 20.08 & 0.468 & 0.411 & \underline{55.85} & 21.58 & 0.556 & 0.338 & \textbf{50.08} \\ 
        DDNM~\cite{wang2022zero} & \textbf{25.23} & \textbf{0.809} & 0.417 & 70.27 & \textbf{22.37} & \textbf{0.689} & 0.596 & 89.41 & 22.28 & 0.677 & 0.629 & 107.9 & 22.02 & 0.705 & 0.415 & 124.4 \\ 
        DDRM~\cite{kawar2022denoising} & 23.98 & 0.763 & 0.378 & 59.58 & 21.29 & 0.653 & 0.554 & 79.71 & 18.35 & 0.553 & 0.677 & 112.7 & 14.42 & 0.401 & 0.892 & 217.5 \\ 
        $\Pi$GDM~\cite{song2022pseudoinverse} & 20.01 & 0.522 & 0.494 & 74.81 & 17.47 & 0.404 & 0.565 & 96.34 & - & - & - & - & 16.05 & 0.313 & 0.708 & 139.9 \\ 
        RED-diff~\cite{mardani2023variational} & 21.52 & 0.604 & 0.633 & 88.91 & 19.64 & 0.539 & 0.767 & 135.43 & 22.85 & 0.651 & 0.363 & 70.15 & 19.45 & 0.622 & 0.534 & 187.7 \\ 
        PnP-ADMM~\cite{chan2016plug} & 24.24 & 0.777 & 0.450 & 82.76 & 21.25 & 0.651 & 0.622 & 144.6 & 23.04 & 0.695 & 0.491 & 86.61 & 8.610 & 0.285 & 0.780 & 256.3 \\ 
        \bottomrule
    \end{tabular}
    }

\label{tab:quantitative_sigma005}
\end{table}

\begin{table}[!t]
    \centering
    \setlength{\tabcolsep}{0.2em}
    
    \caption{Quantitative results of linear tasks including SR $\times4$, SR $\times8$, Gaussian blur, and random inpainting without noise on the ImageNet dataset. The best results are \textbf{bold}, and the second best results are \underline{underlined}.}
    \resizebox{1.0\textwidth}{!}{
    \begin{tabular}{ccccccccccccccccc}
    \toprule 
{} & \multicolumn{4}{c}{\textbf{SR $\times4$}} & \multicolumn{4}{c}{\textbf{SR $\times8$}} &
\multicolumn{4}{c}{\textbf{Gaussian Blur}} &\multicolumn{4}{c}{\textbf{Random Inpainting}} \\
\cmidrule(lr){2-5}
\cmidrule(lr){6-9}
\cmidrule(lr){10-13}
\cmidrule(lr){14-17}
        {\textbf{Method}} & {PSNR $\uparrow$} & {SSIM $\uparrow$} & {LPIPS $\downarrow$} & {FID $\downarrow$}& {PSNR $\uparrow$} & {SSIM $\uparrow$} & {LPIPS $\downarrow$} & {FID $\downarrow$}& {PSNR $\uparrow$} & {SSIM $\uparrow$} & {LPIPS $\downarrow$} & {FID $\downarrow$}& {PSNR $\uparrow$} & {SSIM $\uparrow$} & {LPIPS $\downarrow$} & {FID $\downarrow$} \\ \midrule
        \textbf{DDC} & 25.11 & 0.799 & \textbf{0.137} & \textbf{46.24} & 22.15 & 0.664 & \underline{0.279} & 57.12 & \underline{25.56} & \underline{0.815} & \textbf{0.134} & \textbf{45.99} & \underline{23.54} & \underline{0.770} & \textbf{0.142} & \underline{47.33} \\
        \textbf{DDC w/o LPIPS} & \textbf{25.94} & \textbf{0.828} & 0.274 & 53.21 & \textbf{22.85} & \textbf{0.725} & 0.440 & 67.08 & \textbf{26.62} & \textbf{0.849} & \underline{0.251} & \underline{51.74} & \textbf{23.97} & \textbf{0.784} & 0.223 & 53.35 \\ \midrule
        DPS~\cite{chung2022diffusion} & 24.59 & 0.655 & 0.290 & 48.36 & 20.99 & 0.488 & 0.348 & \underline{49.32} & 20.40 & 0.462 & 0.373 & 52.43 & 22.41 & 0.649 & \underline{0.196} & \textbf{44.94} \\ 
        DDNM~\cite{wang2022zero} & \underline{25.33} & \underline{0.815} & 0.408 & 69.60 & \underline{22.41} & 0.692 & 0.592 & 85.76 & 25.09 & 0.784 & 0.297 & 58.71 & 21.01 & 0.673 & 0.491 & 150.2 \\
        DDRM~\cite{kawar2022denoising} & 25.00 & 0.806 & 0.281 & 53.69 & 21.99 & 0.678 & 0.495 & 72.79 & 24.13 & 0.745 & 0.361 & 63.11 & 14.25 & 0.382 & 0.918 & 220.1 \\ 
        $\Pi$GDM~\cite{song2022pseudoinverse} & 24.66 & 0.802 & \underline{0.179} & \underline{47.37} & 21.24 & 0.659 & \textbf{0.266} & \textbf{48.31} & - & - & - & - & 17.07 & 0.464 & 0.566 & 117.3 \\ 
        RED-diff~\cite{mardani2023variational} & 25.28 & \underline{0.815} & 0.283 & 55.09 & 22.39 & \underline{0.693} & 0.474 & 76.69 & 25.48 & 0.812 & 0.330 & 57.60 & 20.42 & 0.647 & 0.436 & 158.8 \\ 
        PnP-ADMM~\cite{chan2016plug} & 24.65 & 0.787 & 0.398 & 78.50 & 21.43 & 0.656 & 0.579 & 138.1 & 24.30 & 0.782 & 0.520 & 74.00 & 9.020 & 0.301 & 0.765 & 242.1\\ 

        \bottomrule
    \end{tabular}
    }
\label{tab:quantitative_sigma0}
\end{table}

\begin{wraptable}[9]{r}{0.4\textwidth}
    \centering
    \vspace{-0.5cm}
    \caption{Quantative results of JPEG-10 restoration without noise.}
    \setlength{\tabcolsep}{0.2em}
    \resizebox{0.38\textwidth}{!}{
    \begin{tabular}{ccccc}
    \toprule
        {} & \multicolumn{4}{c}{\textbf{JPEG-10}} \\ 
        \cmidrule(lr){2-5}
        \textbf{Method} & PSNR↑ & SSIM↑ & LPIPS↓ & FID↓ \\ \midrule
        \textbf{DDC} & \underline{26.21} & \underline{0.838} & \underline{0.122} & \textbf{45.73} \\
        \textbf{DDC w/o LPIPS} & \textbf{26.84} & \textbf{0.861} & 0.216 & 54.94 \\ \midrule
        $\Pi$GDM~\cite{song2022pseudoinverse} & 24.23 & 0.787 & \textbf{0.108} & 47.52 \\ 
        CDDB~\cite{chung2024direct} & 25.72 & 0.827 & 0.146 & \underline{46.29} \\ 
        \bottomrule
    \end{tabular}
    }
    \label{tab:JPEG-10}
\end{wraptable}

\paragraph{Quality Comparison in nonlinear task} The quantitative results of the JPEG restoration task on the ImageNet dataset are shown in Tab.~\ref{tab:JPEG-10}. Standard DDC has the second-best performance in all three similarity metrics and shows the best realness. Without LPIPS loss, DDC stands out in terms of PSNR and SSIM but its performance is relatively ordinary from the perspective of human perception, which is similar to the linear tasks. $\Pi$GDM has the best LPIPS results but it performs poorly in other metrics. Task-specific trained CDDB is ordinary in the task.

\paragraph{Efficiency} DDC also demonstrates great computational efficiency in Tab.~\ref{tab:steps_and_time} on SR $\times4$ task on 256×256 ImageNet-1k~\cite{deng2009imagenet} dataset. Only 5 sampling steps and 0.77 s on average are required to generate a high-quality image. Other diffusion model-based methods require a runtime ranging from 2.36 s to 220 s because their large inference steps are at least 20 up to 1000. Although our proposed DDC requires an additional training procedure, it takes only about 2 hours for basically one-task model training and 10 hours of training for excellent performance (see Appendix~\ref{sec:appendix_training_process}). Thus even with this time included, our DDC still generates 1000 image solutions faster than most diffusion model-based solvers. Particularly, compared to CDDB, which also requires additional training, our approach saves training time because DDC is a plug-in for the whole diffusion model.

\begin{wraptable}[11]{r}{0.38\textwidth}
    \vspace{-0.4cm}
    \centering
    \caption{Sampling steps and time cost per image on ImageNet dataset.}
    \setlength{\tabcolsep}{0.2em}
    \resizebox{0.28\textwidth}{!}{
    \begin{tabular}{ccc}
    \toprule
        \textbf{Method} &  \textbf{Steps $T$} & \textbf{Time (s)} \\ \midrule
        \textbf{DDC (ours)} & \textbf{5} & \underline{0.77} \\ \midrule
        DPS~\cite{chung2022diffusion} & 1000 & 220 \\ 
        DDNM~\cite{wang2022zero} & 100 & 12.2 \\ 
        DDRM~\cite{kawar2022denoising} & \underline{20} & 2.36 \\ 
        $\Pi$GDM~\cite{song2022pseudoinverse} & 100 & 39.2 \\ 
        RED-diff~\cite{mardani2023variational} & 100 & 12.4 \\ 
        CDDB~\cite{chung2024direct} & \underline{20} & 2.67 \\ 
        PnP-ADMM~\cite{chan2016plug} & - & \textbf{0.68} \\ \bottomrule
    \end{tabular}}
    \label{tab:steps_and_time}
\end{wraptable} 

Two possible reasons may prevent traditional diffusion methods from solving the problem more quickly: (1) the data consistency strength is hard to control, while it may be too strong to deviate from the original diffusion process or too weak to be consistent with the original image. (2) The diffusion process requires enough steps to reconstruct samples. For our DDC approach, the network-guided data consistency can directly update the image with diffusion prior $\hat{\bm{x}}_0$ and measurement $\bm{y}$ to the clean image, and the strength is obtained implicitly in $\Delta_{\phi} \left(\bm{y}, \hat{\bm{x}}_0, t \right)$. Our efficient data consistency step can, in turn, influence the diffusion process, as it yields a high-quality image from the conditional predicted image $\bm{x}_{0,\bm{y}}$ in a short time, so large sampling steps are no longer needed.

\subsection{Results on Robustness}
\label{sec:results_robustness}
\begin{figure}[t]
   \centering
  \begin{minipage}[t]{0.63\linewidth}
    \centering
    \captionof{table}{Quantitative results of the SR $\times4$ and random inpainting tasks on CelebA dataset with Gaussian noise strength $\sigma_y=0.05$. DDC networks are trained on ImageNet and CelebA datasets separately.}
    \setlength{\tabcolsep}{0.2em}
    \resizebox{\textwidth}{!}{
    \begin{tabular}{ccccccccc}
    \toprule
        {} & \multicolumn{4}{c}{\textbf{SR $\times4$}} & \multicolumn{4}{c}{\textbf{Random Inpainting}} \\
        \cmidrule(lr){2-5}
        \cmidrule(lr){6-9}
         {\textbf{Method}} & {PSNR $\uparrow$} & {SSIM $\uparrow$} & {LPIPS $\downarrow$} & {FID $\downarrow$}& {PSNR $\uparrow$} & {SSIM $\uparrow$} & {LPIPS $\downarrow$} & {FID $\downarrow$} \\ \midrule
        DDC (ImageNet) & \underline{29.04} & \underline{0.895} & \underline{0.076} & \underline{18.96} & \textbf{28.18} & \textbf{0.896} & \underline{0.079} & \underline{18.46} \\ 
        DDC (CelebA) & \textbf{29.32} & \textbf{0.901} & \textbf{0.074} & 19.12 & \underline{27.91} & \underline{0.889} & 0.080 & 20.67 \\ \midrule
        DPS~\cite{chung2022diffusion} & 27.88 & 0.871 & 0.085 & \textbf{17.13} & 27.26 & 0.873 & \textbf{0.073} & \textbf{16.67} \\
        DDNM~\cite{wang2022zero} & 24.22& 0.687 & 0.429 & 86.11 & 26.26 & 0.858 & 0.138 & 23.20 \\
        $\Pi$GDM~\cite{song2022pseudoinverse} & 23.46 & 0.657 & 0.353 & 83.27 & 25.66 & 0.793 & 0.183 & 36.94 \\
        RED-diff~\cite{mardani2023variational} & 25.32 & 0.741 & 0.331 & 56.67 & 21.49 & 0.746 & 0.341 & 104.7 \\
        \bottomrule
    \end{tabular}}
    \label{tab:celeba_results}
  \end{minipage}%
  \hspace{0.2cm}
  \begin{minipage}[t]{0.34\linewidth}
    \centering
    \captionof{table}{Comparison between Generalized (G) and Specialized (S) DDC models in Gaussian blur on ImageNet with $\sigma_y = 0.05$.}
    \setlength{\tabcolsep}{0.2em}
    \renewcommand\arraystretch{1.2}
    \resizebox{\textwidth}{!}{
    \begin{tabular}{cccccc}
    \toprule
        $L_{LPIPS}$ & \textbf{Type} & {PSNR $\uparrow$} & {SSIM $\uparrow$} & {LPIPS $\downarrow$} & {FID $\downarrow$} \\ \hline
        \multirow{2}{*}{\checkmark} & G &  23.88 & 0.741 & 0.206 & 49.64 \\ 
         & S & 23.49 & 0.725 & 0.246 & 53.68 \\ \hline
        \multirow{2}{*}{\XSolidBrush} & G & 24.65 & 0.776 & 0.355 & 58.12 \\ 
         & S & 24.62 & 0.775 & 0.380 & 57.84 \\ \bottomrule
    \end{tabular}}
    \label{tab:GvsS}
  \end{minipage}%
  
\end{figure}

\begin{figure}
\includegraphics[width=\textwidth]{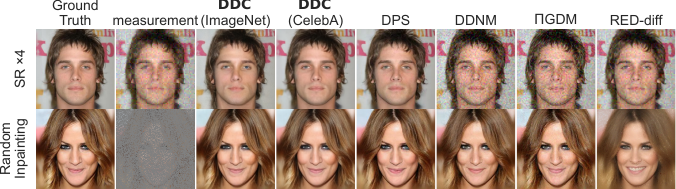}

\caption{Representative results on linear inverse problems with Gaussian noise $\sigma_y = 0.05$ on CelebA dataset. The degradations include SR $\times4$ and random inpainting tasks.}

\label{fig:celeba}
\end{figure}

\paragraph{Robustness across Datasets} To further demonstrate the robustness across datasets, we use the DDC model trained on the ImageNet-1k~\cite{deng2009imagenet} and CelebA~\cite{liu2015faceattributes} dataset separately to solve inverse problems, including SR $\times4$ and 92\% random inpainting tasks with Gaussian noise $\sigma_y=0.05$ on the CelebA dataset. The diffusion network is taken from~\cite{lugmayr2022repaint}. The quantitative and representative results are shown in Tab.~\ref{tab:celeba_results} and Fig.~\ref{fig:celeba}. The performance of the two DDC models is quite close and remarkably favorable to other methods, which indicates that DDC can be effective in grasping the essence of the inverse problems rather than simply fitting to the data. DPS still has the best realness but it performs slightly inferior in terms of similarity.

\begin{wrapfigure}[11]{r}{0.55\textwidth}
    \vspace{-0.3cm}
    \centering
    \setlength{\tabcolsep}{0.2em}
    \resizebox{0.5\textwidth}{!}{
    \includegraphics[width=\textwidth]{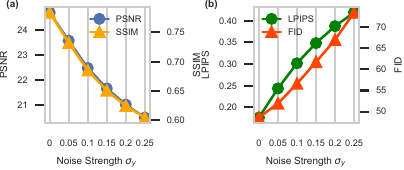}
    }
    \caption{Quantitative results of SR $\times4$ task with different noise strength $\sigma_y$ on ImageNet dataset. (a) PSNR$\uparrow$ and SSIM$\uparrow$, (b) LPIPS$\downarrow$ and FID$\downarrow$.}
    \label{fig:noise_strength_plot}
\end{wrapfigure} 

\begin{figure}
\includegraphics[width=\textwidth]{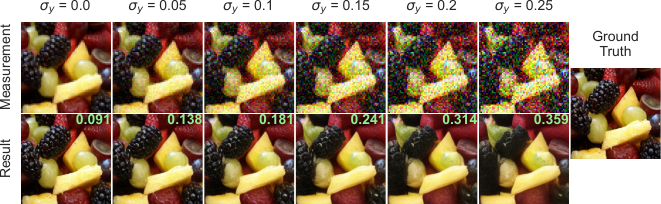}
\caption{Representative results of SR $\times4$ task with different Gaussian noise strength from 0 to 0.25. LPIPS$\downarrow$ metric is labeled on the reconstruction result.}
\label{fig:noise_strength}
\end{figure}

\paragraph{Noise Robustness} DDC has a powerful capacity to solve extremely noisy inverse problems. An SR $\times4$-specific model is trained with random Gaussian noise strength $\sigma_y$ from 0 to 0.25. The quantitative and representative results are shown in Fig.~\ref{fig:noise_strength_plot} and~\ref{fig:noise_strength}. Although the similarity and realness of the results decrease with increasing noise intensity, the reconstructed images at $\sigma_y=0.25$ are still satisfactory, where many image details are still preserved.

\paragraph{Generalized Model vs. Task-specific Model} We evaluate the impact of learning multiple tasks simultaneously on the model. Generalized models perform slightly better than task-specialized ones in the noisy Gaussian blur task as shown in Tab.~\ref{tab:GvsS}. Although the DDC model requires to be fine-tuned when facing new tasks, one generalized model has enough capacity to solve several inverse problems simultaneously, probably because the residual learning is more suitable to be optimized for inverse problems~\cite{zhang2017beyond}. Meanwhile, as can be seen from the comparison of the two experiments with and without LPIPS loss, its inclusion does not affect the two models.
\section{Discussions}
\paragraph{Relationship with Diffusion-based Solvers}Based on previous studies of diffusion model-based inverse problem solvers, DDC puts a step forward by a deep-learning-based data consistency instead of a theoretical-derivation-based one. Previous methods actually have formal commonalities, and their relationship with the diffusion process is worth exploring in depth. DDC uses a network to solve problems more flexibly and directly. Some diffusion-bridge-based solvers~\cite{liu20232, chung2024direct} also utilize networks to learn the connection between measurement and clean data in a modified diffusion process, but they abandon the commonly used diffusion models and are only task-specific solvers. Few studies have embedded other networks within the framework of a diffusion process. Similarly to DDC, CDDM~\cite{chen2024mitigating} also uses a deep learning model to mitigate CT training-sampling discrepancy induced by ADMM data consistency. However, due to the difference between 3D medical and 2D natural images and their problematic starting points, DDC is better suited to the 2D inverse problems.

\paragraph{Relationship with Deep Unrolling} DDC is partially inspired by deep unrolling~\cite{monga2021algorithm} that provides a connection between iterative algorithms and deep learning approaches. For example, PnP-ADMM~\cite{chan2016plug} replaces the denoising step with a neural network denoiser, and ADMM-Net~\cite{yang2016deep} has a series of networks with the same data flow graph as ADMM. Likewise, DDC uses a deep data consistency network as part of a large diffusion framework for solving inverse problems. Its integration with diffusion models is achieved through cleverly designed variational training objectives.

\section{Conclusions}
\label{sec:conclusions}
This paper introduces a novel deep-learning-based data consistency approach DDC for solving inverse problems with the diffusion process. By analyzing the commonalities and drawbacks of previous studies, we propose the residual learning form of DDC and design the training objective to balance the data consistency and realness. Experiments in various inverse problems show excellent outcomes compared to state-of-the-art solvers and demonstrate the high quality of DDC. The fast sampling speed in only 5 inference steps makes it suitable for everyday applications. Moreover, the robustness is shown in tests across datasets, experiments with large noise, and its capacity to solve several tasks in one model. However, to prevent misuse, the training dataset should be carefully selected and existing safeguards for diffusion models could also be useful.

\paragraph{Limitations} One obvious drawback of DDC is that the network should be fine-tuned when facing new inverse problems. Though one DDC model can be trained for several tasks and the training cost is not too large, it still limits the potential wider applications. Future studies may focus on developing a universal network model of data consistency for various types of tasks, which may enable inverse problem solving without requiring a known forward function.

\medskip
\section*{Acknowledgment}
This work was supported by the National Key Research and
Development Program of China under Grant 2021YFC2401301 and
Grant 2021YFC2401300.

\bibliographystyle{plainnat}
\bibliography{main}

\begin{thebibliography}{38}
\providecommand{\natexlab}[1]{#1}
\providecommand{\url}[1]{\texttt{#1}}
\expandafter\ifx\csname urlstyle\endcsname\relax
  \providecommand{\doi}[1]{doi: #1}\else
  \providecommand{\doi}{doi: \begingroup \urlstyle{rm}\Url}\fi

\bibitem[Balke et~al.(2022)Balke, Davis~Rivera, Garcia-Cardona, Majee, McCann, Pfister, and Wohlberg]{balke2022scientific}
Thilo Balke, Fernando Davis~Rivera, Cristina Garcia-Cardona, Soumendu Majee, Michael~Thompson McCann, Luke Pfister, and Brendt~Egon Wohlberg.
\newblock Scientific computational imaging code (scico).
\newblock \emph{Journal of Open Source Software}, 7\penalty0 (LA-UR-22-28555), 2022.

\bibitem[Ballester et~al.(2022)Ballester, Carrillo, Cl{\'e}ment, and Vitoria]{ballester2022analysis}
Coloma Ballester, Hernan Carrillo, Micha{\"e}l Cl{\'e}ment, and Patricia Vitoria.
\newblock Analysis of different losses for deep learning image colorization.
\newblock In \emph{Handbook of Mathematical Models and Algorithms in Computer Vision and Imaging: Mathematical Imaging and Vision}, pages 1--26. Springer, 2022.

\bibitem[Boyd et~al.(2011)Boyd, Parikh, Chu, Peleato, Eckstein, et~al.]{boyd2011distributed}
Stephen Boyd, Neal Parikh, Eric Chu, Borja Peleato, Jonathan Eckstein, et~al.
\newblock Distributed optimization and statistical learning via the alternating direction method of multipliers.
\newblock \emph{Foundations and Trends{\textregistered} in Machine learning}, 3\penalty0 (1):\penalty0 1--122, 2011.

\bibitem[Chan et~al.(2016)Chan, Wang, and Elgendy]{chan2016plug}
Stanley~H Chan, Xiran Wang, and Omar~A Elgendy.
\newblock Plug-and-play admm for image restoration: Fixed-point convergence and applications.
\newblock \emph{IEEE Transactions on Computational Imaging}, 3\penalty0 (1):\penalty0 84--98, 2016.

\bibitem[Chen et~al.(2024)Chen, Hao, Guo, and Xiao]{chen2024mitigating}
Hanyu Chen, Zhixiu Hao, Lin Guo, and Liying Xiao.
\newblock Mitigating data consistency induced discrepancy in cascaded diffusion models for sparse-view ct reconstruction.
\newblock \emph{arXiv preprint arXiv:2403.09355}, 2024.

\bibitem[Chung et~al.(2022{\natexlab{a}})Chung, Kim, Mccann, Klasky, and Ye]{chung2022diffusion}
Hyungjin Chung, Jeongsol Kim, Michael~T Mccann, Marc~L Klasky, and Jong~Chul Ye.
\newblock Diffusion posterior sampling for general noisy inverse problems.
\newblock \emph{arXiv preprint arXiv:2209.14687}, 2022{\natexlab{a}}.

\bibitem[Chung et~al.(2022{\natexlab{b}})Chung, Sim, Ryu, and Ye]{chung2022improving}
Hyungjin Chung, Byeongsu Sim, Dohoon Ryu, and Jong~Chul Ye.
\newblock Improving diffusion models for inverse problems using manifold constraints.
\newblock \emph{Advances in Neural Information Processing Systems}, 35:\penalty0 25683--25696, 2022{\natexlab{b}}.

\bibitem[Chung et~al.(2023)Chung, Lee, and Ye]{chung2023fast}
Hyungjin Chung, Suhyeon Lee, and Jong~Chul Ye.
\newblock Fast diffusion sampler for inverse problems by geometric decomposition.
\newblock \emph{arXiv preprint arXiv:2303.05754}, 2023.

\bibitem[Chung et~al.(2024)Chung, Kim, and Ye]{chung2024direct}
Hyungjin Chung, Jeongsol Kim, and Jong~Chul Ye.
\newblock Direct diffusion bridge using data consistency for inverse problems.
\newblock \emph{Advances in Neural Information Processing Systems}, 36, 2024.

\bibitem[Deng et~al.(2009)Deng, Dong, Socher, Li, Li, and Fei-Fei]{deng2009imagenet}
Jia Deng, Wei Dong, Richard Socher, Li-Jia Li, Kai Li, and Li~Fei-Fei.
\newblock Imagenet: A large-scale hierarchical image database.
\newblock In \emph{2009 IEEE conference on computer vision and pattern recognition}, pages 248--255. Ieee, 2009.

\bibitem[Dhariwal and Nichol(2021)]{dhariwal2021diffusion}
Prafulla Dhariwal and Alexander Nichol.
\newblock Diffusion models beat gans on image synthesis.
\newblock \emph{Advances in neural information processing systems}, 34:\penalty0 8780--8794, 2021.

\bibitem[He et~al.(2023)He, Yan, Luo, Luo, Wang, Du, Chen, Yang, and Zhang]{he2023iterative}
Linchao He, Hongyu Yan, Mengting Luo, Kunming Luo, Wang Wang, Wenchao Du, Hu~Chen, Hongyu Yang, and Yi~Zhang.
\newblock Iterative reconstruction based on latent diffusion model for sparse data reconstruction.
\newblock \emph{arXiv preprint arXiv:2307.12070}, 2023.

\bibitem[Heusel et~al.(2017)Heusel, Ramsauer, Unterthiner, Nessler, and Hochreiter]{heusel2017gans}
Martin Heusel, Hubert Ramsauer, Thomas Unterthiner, Bernhard Nessler, and Sepp Hochreiter.
\newblock Gans trained by a two time-scale update rule converge to a local nash equilibrium.
\newblock \emph{Advances in neural information processing systems}, 30, 2017.

\bibitem[Ho et~al.(2020)Ho, Jain, and Abbeel]{ho2020denoising}
Jonathan Ho, Ajay Jain, and Pieter Abbeel.
\newblock Denoising diffusion probabilistic models.
\newblock \emph{Advances in neural information processing systems}, 33:\penalty0 6840--6851, 2020.

\bibitem[Kadkhodaie and Simoncelli(2021)]{kadkhodaie2021stochastic}
Zahra Kadkhodaie and Eero Simoncelli.
\newblock Stochastic solutions for linear inverse problems using the prior implicit in a denoiser.
\newblock \emph{Advances in Neural Information Processing Systems}, 34:\penalty0 13242--13254, 2021.

\bibitem[Kawar et~al.(2022)Kawar, Elad, Ermon, and Song]{kawar2022denoising}
Bahjat Kawar, Michael Elad, Stefano Ermon, and Jiaming Song.
\newblock Denoising diffusion restoration models.
\newblock \emph{Advances in Neural Information Processing Systems}, 35:\penalty0 23593--23606, 2022.

\bibitem[Li et~al.(2024)Li, Kwon, Alkhouri, Ravishanka, and Qu]{li2024decoupled}
Xiang Li, Soo~Min Kwon, Ismail~R Alkhouri, Saiprasad Ravishanka, and Qing Qu.
\newblock Decoupled data consistency with diffusion purification for image restoration.
\newblock \emph{arXiv preprint arXiv:2403.06054}, 2024.

\bibitem[Liu et~al.(2023)Liu, Vahdat, Huang, Theodorou, Nie, and Anandkumar]{liu20232}
Guan-Horng Liu, Arash Vahdat, De-An Huang, Evangelos~A Theodorou, Weili Nie, and Anima Anandkumar.
\newblock {I$^2$SB: Image-to-Image Schr{\"o}dinger Bridge}.
\newblock \emph{arXiv preprint arXiv:2302.05872}, 2023.

\bibitem[Liu et~al.(2015)Liu, Luo, Wang, and Tang]{liu2015faceattributes}
Ziwei Liu, Ping Luo, Xiaogang Wang, and Xiaoou Tang.
\newblock Deep learning face attributes in the wild.
\newblock In \emph{Proceedings of International Conference on Computer Vision (ICCV)}, December 2015.

\bibitem[Lugmayr et~al.(2022)Lugmayr, Danelljan, Romero, Yu, Timofte, and Van~Gool]{lugmayr2022repaint}
Andreas Lugmayr, Martin Danelljan, Andres Romero, Fisher Yu, Radu Timofte, and Luc Van~Gool.
\newblock Repaint: Inpainting using denoising diffusion probabilistic models.
\newblock In \emph{Proceedings of the IEEE/CVF conference on computer vision and pattern recognition}, pages 11461--11471, 2022.

\bibitem[Mardani et~al.(2023)Mardani, Song, Kautz, and Vahdat]{mardani2023variational}
Morteza Mardani, Jiaming Song, Jan Kautz, and Arash Vahdat.
\newblock A variational perspective on solving inverse problems with diffusion models.
\newblock \emph{arXiv preprint arXiv:2305.04391}, 2023.

\bibitem[Monga et~al.(2021)Monga, Li, and Eldar]{monga2021algorithm}
Vishal Monga, Yuelong Li, and Yonina~C Eldar.
\newblock Algorithm unrolling: Interpretable, efficient deep learning for signal and image processing.
\newblock \emph{IEEE Signal Processing Magazine}, 38\penalty0 (2):\penalty0 18--44, 2021.

\bibitem[Nichol and Dhariwal(2021)]{nichol2021improved}
Alexander~Quinn Nichol and Prafulla Dhariwal.
\newblock Improved denoising diffusion probabilistic models.
\newblock In \emph{International conference on machine learning}, pages 8162--8171. PMLR, 2021.

\bibitem[Odaibo(2019)]{odaibo2019tutorial}
Stephen Odaibo.
\newblock Tutorial: Deriving the standard variational autoencoder (vae) loss function.
\newblock \emph{arXiv preprint arXiv:1907.08956}, 2019.

\bibitem[Robbins(1992)]{robbins1992empirical}
Herbert~E Robbins.
\newblock An empirical bayes approach to statistics.
\newblock In \emph{Breakthroughs in Statistics: Foundations and basic theory}, pages 388--394. Springer, 1992.

\bibitem[Rombach et~al.(2022)Rombach, Blattmann, Lorenz, Esser, and Ommer]{rombach2022high}
Robin Rombach, Andreas Blattmann, Dominik Lorenz, Patrick Esser, and Bj{\"o}rn Ommer.
\newblock High-resolution image synthesis with latent diffusion models.
\newblock In \emph{Proceedings of the IEEE/CVF conference on computer vision and pattern recognition}, pages 10684--10695, 2022.

\bibitem[Ronneberger et~al.(2015)Ronneberger, Fischer, and Brox]{ronneberger2015u}
Olaf Ronneberger, Philipp Fischer, and Thomas Brox.
\newblock U-net: Convolutional networks for biomedical image segmentation.
\newblock In \emph{Medical image computing and computer-assisted intervention--MICCAI 2015: 18th international conference, Munich, Germany, October 5-9, 2015, proceedings, part III 18}, pages 234--241. Springer, 2015.

\bibitem[Rout et~al.(2024)Rout, Raoof, Daras, Caramanis, Dimakis, and Shakkottai]{rout2024solving}
Litu Rout, Negin Raoof, Giannis Daras, Constantine Caramanis, Alex Dimakis, and Sanjay Shakkottai.
\newblock Solving linear inverse problems provably via posterior sampling with latent diffusion models.
\newblock \emph{Advances in Neural Information Processing Systems}, 36, 2024.

\bibitem[Sohl-Dickstein et~al.(2015)Sohl-Dickstein, Weiss, Maheswaranathan, and Ganguli]{sohl2015deep}
Jascha Sohl-Dickstein, Eric Weiss, Niru Maheswaranathan, and Surya Ganguli.
\newblock Deep unsupervised learning using nonequilibrium thermodynamics.
\newblock In \emph{International conference on machine learning}, pages 2256--2265. PMLR, 2015.

\bibitem[Song et~al.(2023)Song, Kwon, Zhang, Hu, Qu, and Shen]{song2023solving}
Bowen Song, Soo~Min Kwon, Zecheng Zhang, Xinyu Hu, Qing Qu, and Liyue Shen.
\newblock Solving inverse problems with latent diffusion models via hard data consistency.
\newblock \emph{arXiv preprint arXiv:2307.08123}, 2023.

\bibitem[Song et~al.(2022)Song, Vahdat, Mardani, and Kautz]{song2022pseudoinverse}
Jiaming Song, Arash Vahdat, Morteza Mardani, and Jan Kautz.
\newblock Pseudoinverse-guided diffusion models for inverse problems.
\newblock In \emph{International Conference on Learning Representations}, 2022.

\bibitem[Song et~al.(2020)Song, Sohl-Dickstein, Kingma, Kumar, Ermon, and Poole]{song2020score}
Yang Song, Jascha Sohl-Dickstein, Diederik~P Kingma, Abhishek Kumar, Stefano Ermon, and Ben Poole.
\newblock Score-based generative modeling through stochastic differential equations.
\newblock \emph{arXiv preprint arXiv:2011.13456}, 2020.

\bibitem[Su et~al.(2022)Su, Xu, and Yin]{su2022survey}
Jingwen Su, Boyan Xu, and Hujun Yin.
\newblock A survey of deep learning approaches to image restoration.
\newblock \emph{Neurocomputing}, 487:\penalty0 46--65, 2022.

\bibitem[von Platen et~al.(2022)von Platen, Patil, Lozhkov, Cuenca, Lambert, Rasul, Davaadorj, Nair, Paul, Berman, Xu, Liu, and Wolf]{von-platen-etal-2022-diffusers}
Patrick von Platen, Suraj Patil, Anton Lozhkov, Pedro Cuenca, Nathan Lambert, Kashif Rasul, Mishig Davaadorj, Dhruv Nair, Sayak Paul, William Berman, Yiyi Xu, Steven Liu, and Thomas Wolf.
\newblock Diffusers: State-of-the-art diffusion models.
\newblock \url{https://github.com/huggingface/diffusers}, 2022.

\bibitem[Wang et~al.(2022)Wang, Yu, and Zhang]{wang2022zero}
Yinhuai Wang, Jiwen Yu, and Jian Zhang.
\newblock Zero-shot image restoration using denoising diffusion null-space model.
\newblock \emph{arXiv preprint arXiv:2212.00490}, 2022.

\bibitem[Yang et~al.(2016)Yang, Sun, Li, and Xu]{yang2016deep}
Yan Yang, Jian Sun, Huibin Li, and Zongben Xu.
\newblock Deep admm-net for compressive sensing mri.
\newblock In \emph{Proceedings of the 30th international conference on neural information processing systems}, pages 10--18, 2016.

\bibitem[Zhang et~al.(2017)Zhang, Zuo, Chen, Meng, and Zhang]{zhang2017beyond}
Kai Zhang, Wangmeng Zuo, Yunjin Chen, Deyu Meng, and Lei Zhang.
\newblock Beyond a gaussian denoiser: Residual learning of deep cnn for image denoising.
\newblock \emph{IEEE transactions on image processing}, 26\penalty0 (7):\penalty0 3142--3155, 2017.

\bibitem[Zhang et~al.(2018)Zhang, Isola, Efros, Shechtman, and Wang]{zhang2018unreasonable}
Richard Zhang, Phillip Isola, Alexei~A Efros, Eli Shechtman, and Oliver Wang.
\newblock The unreasonable effectiveness of deep features as a perceptual metric.
\newblock In \emph{Proceedings of the IEEE conference on computer vision and pattern recognition}, pages 586--595, 2018.

\end{thebibliography}

\clearpage
\appendix
\section{Additional Method Details}
\label{sec:additional method}
\subsection{Commonalities of Existing Methods}
\label{sec:appendix_commonalities}

Most of the diffusion model-based solvers for inverse problems share commonalities in their mathematical forms. To update the data consistency step, some methods directly modify noisy image $ \bm{x}_{t, \bm{y}} = \bm{x}_t - \Delta$, others modify the predicted clean image $\hat{\bm{x}}_{0, \bm{y}} = \hat{\bm{x}}_0 - \Delta$. Actually, these two ways are equivalent in form. For example, in DDPM~\cite{ho2020denoising} sampling, the relationship can be written as:
\begin{equation}
    \bm{x}_{t-1, \bm{y}} = \frac{\sqrt{\bar{\alpha}_t}\beta_t}{1 - \bar{\alpha}_t} \hat{\bm{x}}_{0, \bm{y}} + \frac{\sqrt{\alpha_t} (1 - \bar{\alpha}_{t-1})}{1 - \bar{\alpha}_t} \bm{x}_t + \sigma_t \bm{\epsilon}, \bm{\epsilon} \sim \mathcal{N} (\bm{0}, \bm{I}),
\end{equation}
where $\sigma_t$ can be chosen as either fixed size $\beta_t$ and $\frac{1 - \bar{\alpha}_{t-1}}{1 - \bar{\alpha}_t} \beta_t$, or learned variance $\bm{\Sigma}_{\theta} (\bm{x}_t, t)$. Here we introduce three representative approaches.

\paragraph{DPS} DPS~\cite{chung2022diffusion} updates the data consistency in the noisy image $\bm{x}_t$. It provides a tractable approximation  $p(\bm{y} | \bm{x}_t) \simeq p(\bm{y} | \hat{\bm{x}}_0)$, where $\hat{\bm{x}}_0$ is predicted by the Tweedie's approach~\cite{robbins1992empirical}. Assuming the noise is Gaussian, the gradient of the log-likelihood is approximated as $\nabla_{\bm{x}_t} \log p (\bm{y} | \bm{x}_t) \simeq \zeta_t \nabla_{\bm{x}_t} \Vert \bm{y} - \mathcal{A}(\hat{\bm{x}}_0) \Vert_2^2$. By leveraging Eq.~\ref{eq:rewrite_score}, we have the data consistency step
\begin{equation}
    \bm{x}_{t-1} = \bm{x}^{\prime}_{t-1} - \zeta_t \nabla_{\bm{x}_t} \Vert \bm{y} - \mathcal{A}(\hat{\bm{x}}_0) \Vert_2^2,
\end{equation}
where $\bm{x}^{\prime}_{t-1}$ is the ~\cite{song2022pseudoinverse}estimation by the reverse diffusion process. $\Pi$GDM~\cite{song2022pseudoinverse}, PSLD~\cite{rout2024solving} and HGU~\cite{he2023iterative} have similar update forms.

\paragraph{DDNM} DDNM~\cite{wang2022zero} updates data consistency in the predicted clean image $\hat{\bm{x}}_0$. The range-null space decomposition is useful to make the range-space satisfy $\bm{A}\hat{\bm{x}}_0 = \bm{y}$ and leave the null-space $(\bm{I} - \bm{A}^{\dagger}\bm{A}) \hat{\bm{x}}_0$ unchanged. The data consistency step can be written as:
\begin{equation}
    \hat{\bm{x}}_{0, \bm{y}} = \hat{\bm{x}}_0 - \bm{\Sigma}_t \bm{A}^{\dagger} (\bm{A} \hat{\bm{x}}_0 - \bm{y}),
\end{equation}
where $\bm{\Sigma}_t$ is used to scale the range-space correction.

\paragraph{DDS} DDS~\cite{chung2023fast} also updates data consistency in the clean data $\hat{\bm{x}}_0$, but it uses the complex optimization method conjugate gradient (CG). Although this method is a k-step iterative method, its update to $\hat{\bm{x}}_0$ can be summarized as:
\begin{equation}
    \hat{\bm{x}}_{0, \bm{y}} = \hat{\bm{x}}_0 + \alpha_k \bm{p}_k,
\end{equation}
where $\alpha_k$ and $\bm{p}_k$ are parameters related to $\bm{y}$, $\bm{A}$, and $\hat{\bm{x}}_0$. Resample~\cite{song2023solving} has a similar data consistency update form.

In summary, these methods update data consistency in similar forms $\hat{\bm{x}}_{0, \bm{y}} = \hat{\bm{x}}_0 - \Delta$, and the $\Delta$ is related to $\bm{y}$, $\bm{A}$, $\hat{\bm{x}}_0$, and noise schedules $\beta_t$. Accordingly, we propose a deep learning-based DDC.

\subsection{Detailed Derivations of DDC}
\label{sec:appendix_variational_bound_detailed}

The conditional posterior $p(\hat{\bm{x}}_{0, \bm{y}}|\bm{y})$ is expected to be maximized when training the data consistency network $\phi$. Since $D_{KL} (q(\bm{x}_t | \bm{x}_0, \bm{y}) \Vert p(\bm{x}_t | \hat{\bm{x}}_{0, \bm{y}}, \bm{y})) \geq 0$, the upper bound on negative log-likelihood can be written as:
\begin{align*}
    \mathbb{E}_{\bm{x}_0, \bm{y}} \left[-\log p(\hat{\bm{x}}_{0, \bm{y}}|\bm{y}) \right] &\leq \mathbb{E}_{\bm{x}_0, \bm{y}} \left[ -\log p(\hat{\bm{x}}_{0, \bm{y}}|\bm{y}) \right] + D_{KL} (q(\bm{x}_t | \bm{x}_0 , \bm{y}) \Vert p(\bm{x}_t | \hat{\bm{x}}_{0, \bm{y}} , \bm{y})) \\
    &= \mathbb{E}_{\bm{x}_0, \bm{y}} \left[ -\log p(\hat{\bm{x}}_{0, \bm{y}}|\bm{y}) \right] - \mathbb{E}_{\bm{x}_0, \bm{y}, q(\bm{x}_t | \bm{x}_0 , \bm{y})} \left [\log \frac{p(\bm{x}_t | \hat{\bm{x}}_{0, \bm{y}} , \bm{y})}{q(\bm{x}_t | \bm{x}_0 , \bm{y})} \right ] \\
    &= - \mathbb{E}_{\bm{x}_0, \bm{y}, q(\bm{x}_t | \bm{x}_0 , \bm{y})} \left [\log p(\hat{\bm{x}}_{0, \bm{y}}|\bm{y}) + \log \frac{p(\hat{\bm{x}}_{0, \bm{y}} | \bm{x}_t , \bm{y}) p(\bm{x}_t | \bm{y})}{q(\bm{x}_t | \bm{x}_0 , \bm{y}) p(\hat{\bm{x}}_{0, \bm{y}} | \bm{y}) } \right ] \\
    &= - \mathbb{E}_{\bm{x}_0, \bm{y}, q(\bm{x}_t | \bm{x}_0 , \bm{y})} \left [\log p(\hat{\bm{x}}_{0, \bm{y}} | \bm{x}_t , \bm{y}) + \log \frac{p(\bm{x}_t | \bm{y})}{q(\bm{x}_t | \bm{x}_0 , \bm{y}) } \right ] \\
    &= - \mathbb{E}_{\bm{x}_0, \bm{y}, q(\bm{x}_t | \bm{x}_0 , \bm{y})} \left [\log p(\hat{\bm{x}}_{0, \bm{y}} | \bm{x}_t , \bm{y}) \right ] + D_{KL} (q(\bm{x}_t | \bm{x}_0 , \bm{y}) \Vert p(\bm{x}_t | \bm{y})).
\end{align*}
The variational bound can be used as the training objective with the expectation for all timesteps $t$ as Eq.~\ref{eq:training_objective}:
\begin{equation*}
    L_{\phi} = \underbrace{- \mathbb{E}_{\bm{x}_0, t, \bm{x}_t, \bm{y}} \left [\log p(\hat{\bm{x}}_{0, \bm{y}} | \bm{x}_t , \bm{y}) \right ]}_{L_{recon}} + \underbrace{\mathbb{E}_{\bm{x}_0, t, \bm{x}_t, \bm{y}}D_{KL} (q(\bm{x}_t | \bm{x}_0 , \bm{y}) \Vert p(\bm{x}_t | \bm{y}))}_{L_{KL}}.
\end{equation*}

For KL loss, under the assumption that $\bm{x}_t$ is independent of the measurement $\bm{y}$, we can obtain $q(\bm{x}_t | \bm{x}_0 , \bm{y}) = q(\bm{x}_t | \bm{x}_0) \sim \mathcal{N} (\sqrt{\alpha_t \bm{x}_0}, \sqrt{1-\alpha_t}\bm{I})$. If we assume that the conditional reverse process is also Gaussian $p(\bm{x}_t | \bm{y}) \sim \mathcal{N} (\bm{\mu}_t, \bm{\sigma}_t)$, the KL diverge is a closed-form function.

Generally speaking, for two one-dimensional Gaussian distributions $q(x) \sim \mathcal{N} (\mu_1, \sigma_1)$ and $p(x) \sim \mathcal{N} (\mu_2, \sigma_2)$, the KL divergence can be written as~\cite{odaibo2019tutorial}:
\begin{align}
    D_{KL} \left( q(x) \Vert p(x)\right) &= \int {q(x) \log \frac{q(x)}{p(x)}} dx  \notag \\
    &= \int {\frac{1}{\sqrt{2\pi \sigma_1^2}} \exp{ \left (-\frac{(x-\mu_1)^2}{2\sigma_1^2} \right)} \log \frac{\frac{1}{\sqrt{2\pi \sigma_1^2}} \exp{ \left (-\frac{(x-\mu_1)^2}{2\sigma_1^2} \right)}}{\frac{1}{\sqrt{2\pi \sigma_2^2}} \exp{ \left (-\frac{(x-\mu_2)^2}{2\sigma_2^2} \right)}} dx } \notag \\
    &= \log{\frac{\sigma_2}{\sigma_1}} + \frac{1}{2 \sigma_2} \left( \sigma_1^2 + \left( \mu_1 - \mu_2 \right)^2 \right) - \frac{1}{2}.
\label{eq:kl_gaussian}
\end{align}
Although the noisy images are multidimensional Gaussian distributions, their dimensions can be considered separately for calculation, given that the covariance matrix $\sqrt{1 - \alpha_t} \bm{I}$ is a scaled identity one. Therefore, in the implementation, we can directly calculate the mean $\mu_t$ and variance $\sigma_t$ of the reverse noisy images, and take them into Eq.~\ref{eq:kl_gaussian} to get the KL loss:
\begin{equation*}
    L_{KL} = \frac{1}{2} \left ( \log \frac{\sigma_t^2}{1 - {{\bar \alpha }_t}} + \frac{1 - {{\bar \alpha }_t} + \left(\sqrt {{{\bar \alpha }_t}}x_0 - \mu_t \right)^2}{\sigma_t^2} -1\right ).
\end{equation*}

\section{Experimental Details}
\label{sec: appendix_experimental_details}

\subsection{DDC Details}
\paragraph{Network Architecture} Similar to the diffusion process, the DDC network $\phi$ is based on U-Net~\cite{ronneberger2015u} using \textit{Diffusers}~\cite{von-platen-etal-2022-diffusers} package. The number of input channels is set to 6, where 3 channels are for the noisy image $\bm{x}_t$ and the other 3 channels are for the measurement $\bm{y}$. The timestep $t$ is positionally embedded to feed the network. The number of output channels for each U-Net block is $[128, 128, 256, 256, 512, 512]$. The second last block is a ResNet block with spatial self-attention and the other layers are regular ResNet blocks. The model has 113.7 million parameters in total.

\paragraph{Network Training}The DDC training procedure is on one single NVIDIA GeForce RTX 4090 GPU and Intel Core i9-12900KF Processor. The generalized model for ImageNet is trained at batch size 2, gradient accumulation step 4, and training steps about 160000. During training, six tasks including SR $\times4$, SR $\times8$, Gaussian blur, 92\% random inpainting, JPEG-10, and denoising are randomly selected with a random Gaussian noise $\sigma_y$ from 0 to 0.1. The training speed is 1.44 seconds per gradient step and the total time is 64.0 hours. The exponential moving average (EMA) model parameters are used as the final DDC model, with a decay factor of 0.9999. The weight of loss is $w_1 = 1$, $w_2 = 0.1$, and $w_3 = 1 \times 10^{-3}$, which is selected by grid search.

\paragraph{Sampling Details} The way to reduce the sampling steps of DDC is the same as~\cite{nichol2021improved} because the diffusion model of ImageNet learns $\bm{\Sigma}_\theta (\bm{x}_t, t)$~\cite{dhariwal2021diffusion}. For an arbitrary subsequence $S$ of timesteps, $\bar{\alpha}_t$ remains the same, and the corresponding sampling variances become:
\begin{equation}
    \beta_{S_t} = 1 - \frac{\bar{\alpha}_{S_t}}{\bar{\alpha}_{S_{t-1}}}, \Tilde{\beta}_{S_t} = \frac{1 - \bar{\alpha}_{S_{t-1}}}{1 - \bar{\alpha}_{S_t}} \beta_{S_t}.
\end{equation}
Then, we can turn the model output $\bm{v}$ into variances as follows:
\begin{equation}
    \bm{\Sigma}_\theta (\bm{x}_t, t) = \exp [ \bm{v}\log \beta_{S_t} + (1-\bm{v}) \log \Tilde{\beta}_{S_t}].
\end{equation}
The one-step unconditional image generation is:
\begin{equation}
    \bm{x}_{S_{t-1}} = \frac{1}{\sqrt{\alpha_{S_t}}} \left( \bm{x}_{S_t} - \frac{\beta_{S_t}}{\sqrt{1 - \bar{\alpha}_{S_t}}} \bm{\epsilon}_\theta \left( \bm{x}_t, t \right) \right) + \bm{\Sigma}_\theta (\bm{x}_t, t) \odot \bm{z}, \bm{z} \sim \mathcal{N} (\bm{0}, \bm{I}).
\end{equation}

For DDC sampling, Eq.~\ref{eq:ddc_update} is used to update data consistency at the predicted clean image $\bm{x}_0$. We can write its equivalent form to update noisy image $\bm{x}_{S_t}$. Based on the posterior $q(\bm{x}_{S_{t-1}} | \bm{x}_{S_t}, \bm{x}_{0}) = \mathcal{N} (\Tilde{\bm{\mu}}_{S_t}, \Tilde{\beta}_{S_t}\bm{I})$ defined as follows:
\begin{equation}
    \Tilde{\bm{\mu}}_{S_t} (\bm{x}_{S_t}, \bm{x}_0) = \frac{\sqrt{\bar{\alpha}_{S_{t-1}}} \beta_{S_t}}{1 - \bar{\alpha}_{S_t}} \bm{x}_0 + \frac{\sqrt{\alpha_{S_t}} (1 - \bar{\alpha}_{S_{t-1}} )}{1 - \bar{\alpha}_{S_t}} \bm{x}_{S_t},
\end{equation}
the DDC update step is
\begin{equation}
    \bm{x}_{S_{t-1}, \bm{y}} = \bm{x}_{S_{t-1}} - \frac{\sqrt{\bar{\alpha}_{S_{t-1}}} \beta_{S_t}}{1 - \bar{\alpha}_{S_t}} \Delta_{\phi} \left(\bm{y}, \hat{\bm{x}}_0, S_t \right).
\end{equation}

\subsection{Details of Comparison Methods}
The diffusion models used for the comparison methods are the same as DDC, which are taken from~\cite{dhariwal2021diffusion} for ImageNet and from~\cite{lugmayr2022repaint} for the CelebA dataset.

\paragraph{DPS} The number of inference steps is set to 1000. The step size $\zeta$ in $\zeta / \Vert \bm{y} - \mathcal{A}(\hat{\bm{x}}_0 (\bm{x}_t)) \Vert$ is 0.4 for Gaussian blur task on ImageNet dataset, 1.0 for all the other tasks on ImageNet dataset, and 1.0 for all the tasks on CelebA dataset.

\paragraph{DDNM} The number of inference steps is 100 and the stochasticity of the process $\eta$ is 0.85. The time travel trick is not used.

\paragraph{DDRM} The number of inference steps is 20. $\eta$ and $\eta_b$ are set to 0.85 and 1.0.

\paragraph{$\Pi$GDM} The number of inference steps is 100 and $\eta = 1.0$ as suggested for all the problems.

\paragraph{RED-diff} The number of inference steps is 100. The optimizer learning rate $lr$ is 0.5, and $\lambda$ is 0.25.

\paragraph{CDDB} CDDB is only used for the SR $\times4$ and JPEG-10 task without noise because the provided pre-trained checkpoints only include the cases. The step size $c$ is 1.0 for the SR $\times4$ problem and 0.5 for the JPEG-10 task.

\paragraph{PnP-ADMM} $scico$ package~\cite{balke2022scientific} is used to reconstruct the images. The parameters are set as $\rho = 0.05$ and $maxiter = 10$. The DnCNN~\cite{zhang2017beyond} is chosen as the denoiser.

\section{Additional Results}
\subsection{Inference Steps}
\begin{figure}
    \centering
    \includegraphics[width=0.6\textwidth]{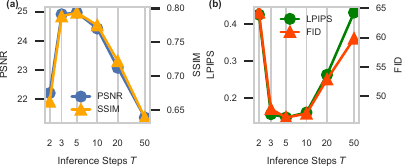}
    \caption{Quantitative results of SR $\times4$ task with different inference steps $T$ on ImageNet dataset. (a) PSNR$\uparrow$ and SSIM$\uparrow$, (b) LPIPS$\downarrow$ and FID$\downarrow$.}
    \label{fig:inference_steps}
\end{figure}

The performance of DDC with respect to the change in inference steps $T$ is shown in Fig.~\ref{fig:inference_steps}. The experiment is taken on the SR $\times4$ task without noise. DDC performs best at $T=5$, the likely reason is that the diffusion model plays a little role when steps are too small ($T=2$), and the majority of the restoration is provided by DDC update Eq.~\ref{eq:ddc_update}. As the steps increase, the image quality of results decreases because multiple data consistency updates may bring the image too close to the measurement $\bm{y}$.

\subsection{Training Process}
\label{sec:appendix_training_process}
\begin{figure}
    \centering
    \includegraphics[width=0.7\textwidth]{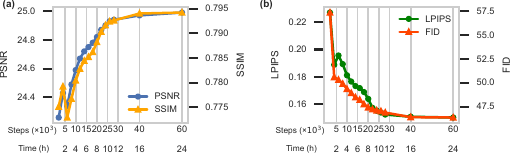}
    \caption{Quantitative results of SR $\times4$ task with different training steps on ImageNet dataset. (a) PSNR$\uparrow$ and SSIM$\uparrow$, (b) LPIPS$\downarrow$ and FID$\downarrow$.}
    \label{fig:training_process}
\end{figure}

Fig.~\ref{fig:training_process} shows the variation trend of metrics with the training steps on 1000 validation images from ImageNet. The model is trained for the single SR $\times4$ task without noise. Consistent with the training process in Sec.~\ref{sec: appendix_experimental_details}, one training step includes 4 gradient accumulation steps with batch size $=2$. The EMA model is used to sample the output. For the task-specific DDC model, a model trained for 2 hours (5000 steps) was sufficient to achieve a basic reconstruction. After more than 10 hours of training (25000 steps), the model's improvement in results was largely insignificant.

\subsection{Additional Visual Results}
\begin{figure}
    \centering
    \includegraphics[width=0.9\textwidth]{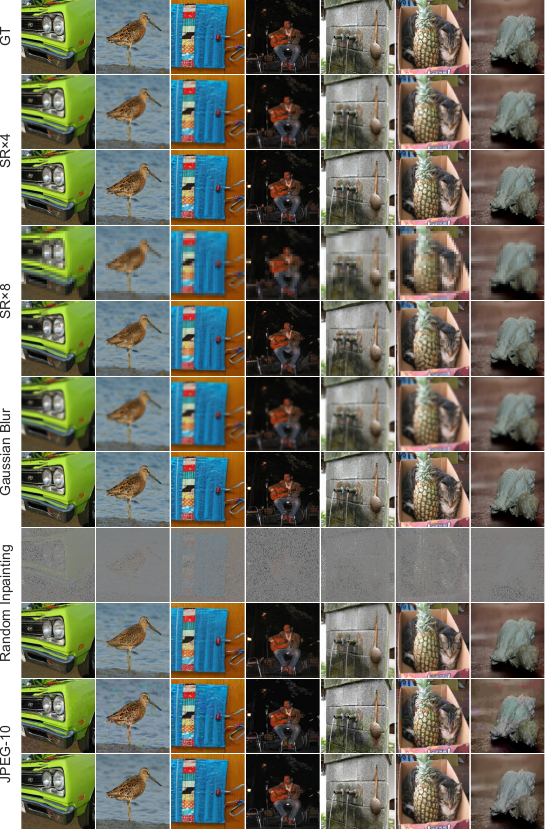}
    \caption{Representative results of standard DDC on different inverse problems without noise on the ImageNet dataset. Ground truth (GT) images are shown in the first row. For each task, the top row is the measurement and the bottom row is the result.}
    \label{fig:appendix_sigma0_imagenet}
\end{figure}

\begin{figure}
    \centering
    \includegraphics[width=0.9\textwidth]{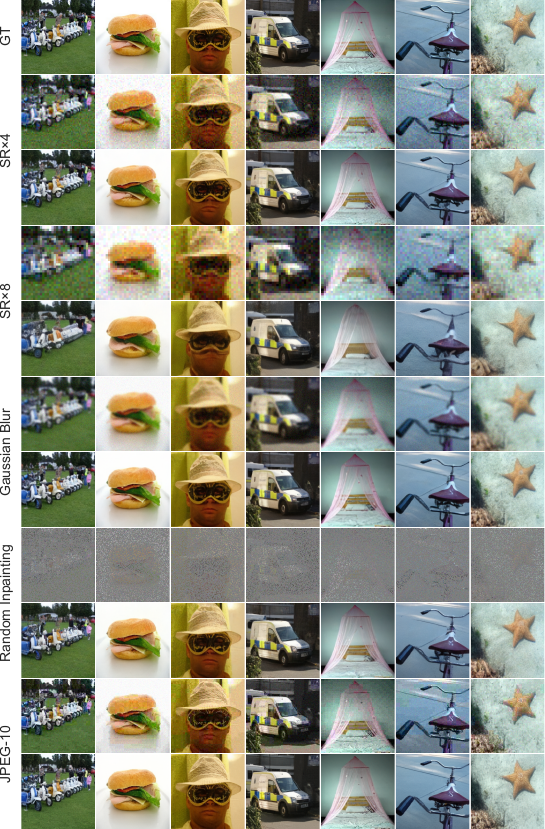}
    \caption{Representative results of standard DDC on different inverse problems with Gaussian noise $\sigma_y = 0.05$ on the ImageNet dataset.}
    \label{fig:appendix_sigma005_imagenet}
\end{figure}

\begin{figure}
    \centering
    \includegraphics[width=0.9\textwidth]{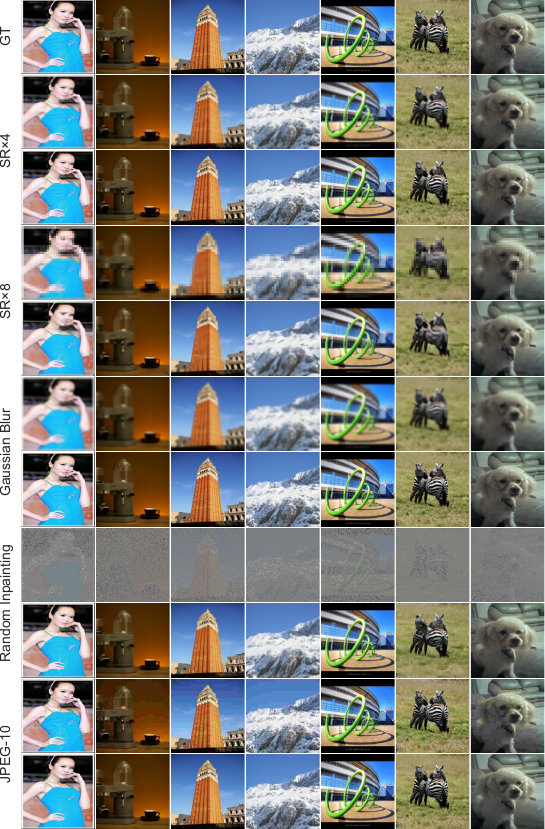}
    \caption{Representative results of DDC without LPIPS loss on different inverse problems without noise on the ImageNet dataset.}
    \label{fig:appendix_sigma0_wolpips_imagenet}
\end{figure}

\begin{figure}
    \centering
    \includegraphics[width=0.9\textwidth]{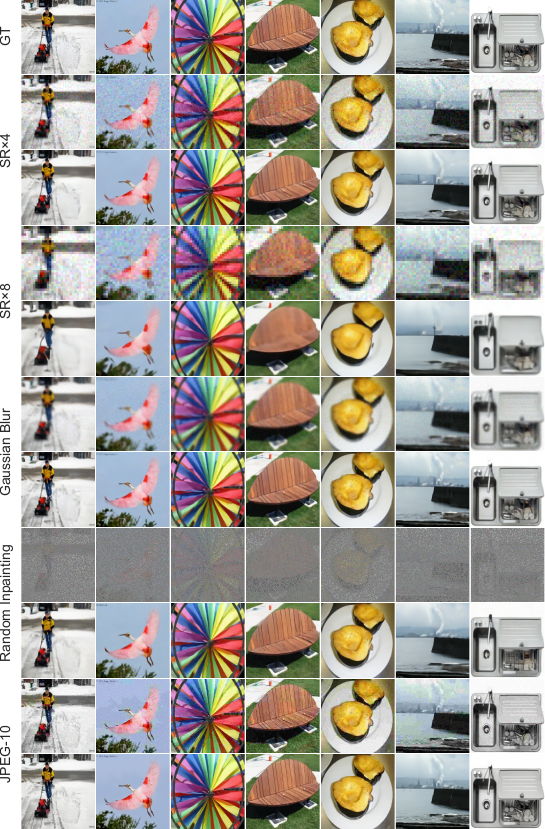}
    \caption{Representative results of DDC without LPIPS loss on different inverse problems with Gaussian noise $\sigma_y = 0.05$ on the ImageNet dataset.}
    \label{fig:appendix_sigma005_wolpips_imagenet}
\end{figure}

\begin{figure}
    \centering
    \includegraphics[width=0.9\textwidth]{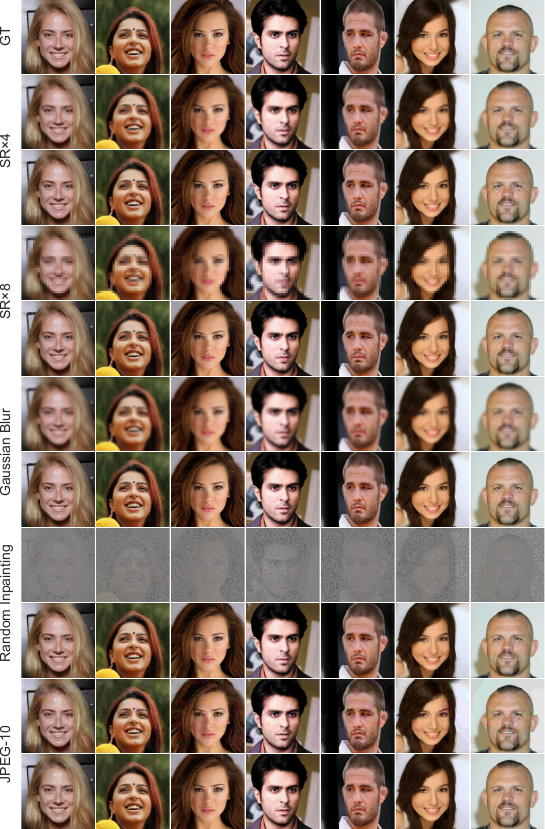}
    \caption{Representative results of standard DDC on different inverse problems without noise on the CelebA dataset.}
    \label{fig:appendix_sigma0_celeba}
\end{figure}

\begin{figure}
    \centering
    \includegraphics[width=0.9\textwidth]{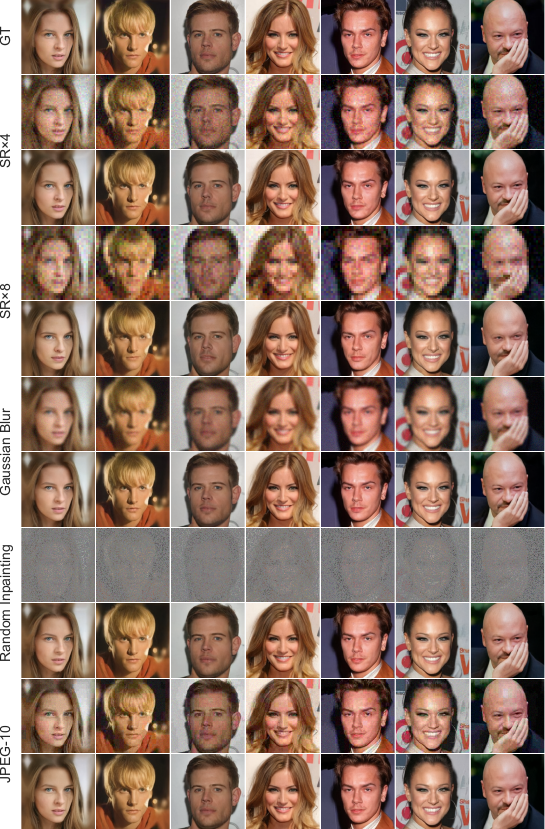}
    \caption{Representative results of standard DDC on different inverse problems with Gaussian noise $\sigma_y = 0.05$ on the CelebA dataset.}
    \label{fig:appendix_sigma005_celeba}
\end{figure}

We present more representative results of our proposed DDC here. Fig.~\ref{fig:appendix_sigma0_imagenet} and Fig.~\ref{fig:appendix_sigma005_imagenet} show some typical results of the standard DDC model (trained with LPIPS loss) on the ImageNet dataset with Gaussian noise $\sigma_y =$ 0 and 0.05 respectively. Fig.~\ref{fig:appendix_sigma0_wolpips_imagenet} and Fig.~\ref{fig:appendix_sigma005_wolpips_imagenet} are the results of the DDC model trained without LPIPS loss on the ImageNet dataset with Gaussian noise $\sigma_y =$ 0 and 0.05 respectively. Fig.\ref{fig:appendix_sigma0_celeba} and Fig.~\ref{fig:appendix_sigma005_celeba} are results of the standard DDC model which trained on the ImageNet dataset and sampled on the CelebA dataset with Gaussian noise $\sigma_y =$ 0 and 0.05 respectively.

\end{document}